\documentclass{article}

\usepackage[english]{babel}

\usepackage[letterpaper,top=2cm,bottom=2cm,left=3cm,right=3cm,marginparwidth=1.75cm]{geometry}

\usepackage{amsmath}
\usepackage{amsthm}
\usepackage[ruled]{algorithm2e}
\usepackage{algpseudocode}
\usepackage{microtype}
\usepackage{graphicx}
\usepackage[usenames,dvipsnames]{xcolor}
\usepackage[colorlinks = true,
            linkcolor = Blue,
            urlcolor  = Blue,
            citecolor = Blue,
            anchorcolor =  Blue]{hyperref}
\usepackage{mathtools}
\usepackage{amsthm}
\usepackage{bibentry}
\usepackage{array}
\usepackage{verbatim}
\usepackage{enumitem}
\usepackage{amssymb}
\usepackage{subcaption}
\usepackage{booktabs} 
\usepackage{mathtools}
\usepackage[numbers]{natbib}
\usepackage{bbm}
\usepackage[mode=buildnew]{standalone}
\usepackage{tikz}
\usetikzlibrary{external}
\newcommand{\data}{\mathcal{D}}
\newcommand{\affiliation}[1]{%
\begingroup%
\let\thefootnote\relax%
\footnotetext{#1}%
\endgroup%
}

\theoremstyle{definition}
\newtheorem{example}{Example}[section]

\title{GFlowNets for AI-Driven Scientific Discovery}
\author{Moksh Jain$^{1,2}$, Tristan Deleu$^{1,2}$, Jason Hartford$^{1,2}$, Cheng-Hao Liu$^{3,2}$,\\Alex Hernandez-Garcia$^{1,2}$, Yoshua Bengio$^{1,2,4}$}
\date{}

\begin{document}
\maketitle
\affiliation{$^1$ Universit\'e de Montr\'eal,$^2$ Mila Quebec AI Institute, $^3$ McGill University, $^4$ CIFAR Fellow \& IVADO}
\begin{abstract}

Tackling the most pressing problems for humanity, such as the climate crisis and the threat of global pandemics, requires accelerating the pace of scientific discovery. While science has traditionally relied on trial and error and even serendipity to a large extent, the last few decades have seen a surge of data-driven scientific discoveries. However, in order to truly leverage large-scale data sets and high-throughput experimental setups, machine learning methods will need to be further improved and better integrated in the scientific discovery pipeline. A key challenge for current machine learning methods in this context is the efficient exploration of very large search spaces, which requires techniques for estimating reducible (epistemic) uncertainty and generating sets of diverse and informative experiments to perform. This motivated a new probabilistic machine learning framework called GFlowNets, which can be applied in the modeling, hypotheses generation and experimental design stages of the experimental science loop. GFlowNets learn to sample from a distribution given indirectly by a reward function corresponding to an unnormalized probability, which enables sampling diverse, high-reward candidates. GFlowNets can also be used to form efficient and amortized Bayesian posterior estimators for causal models conditioned on the already acquired experimental data. Having such posterior models can then provide estimators of epistemic uncertainty and information gain that can drive an experimental design policy. Altogether, here we will argue that GFlowNets can become a valuable tool for AI-driven scientific discovery, especially in scenarios of very large candidate spaces where we have access to cheap but inaccurate measurements or too expensive but accurate measurements. This is a common setting in the context of drug and material discovery, which we use as examples throughout the paper.
\end{abstract}

\clearpage
\tableofcontents
\clearpage
\section{Introduction}
\label{sec:intro}
The climate crisis, antibiotic resistance and the prospect of new pandemics are some of the biggest threats to humanity, posing immense risks to global health and food security. One important common aspect to all these threats and others is that significant new scientific discoveries are required to mitigate them. According to the 2022 report by the Intergovernmental Panel on Climate Change (IPCC)~\citep{ipcc2022}, limiting global warming will require the adoption of alternative fuels, as well as improvements in the efficiency of energy production and material synthesis. The discovery of new materials, such as electrocatalysts that improve the energy efficiency of chemical reactions, can therefore play a crucial role in such a transition. Correspondingly, growing risks of antimicrobial resistance and pandemics make it essential to accelerate the pipeline for discovery of new drugs. Consequently, the well-being of our societies will strongly depend on the pace of our scientific discoveries.

\begin{figure}[t]
    \centering
    \includegraphics[width=0.5\textwidth]{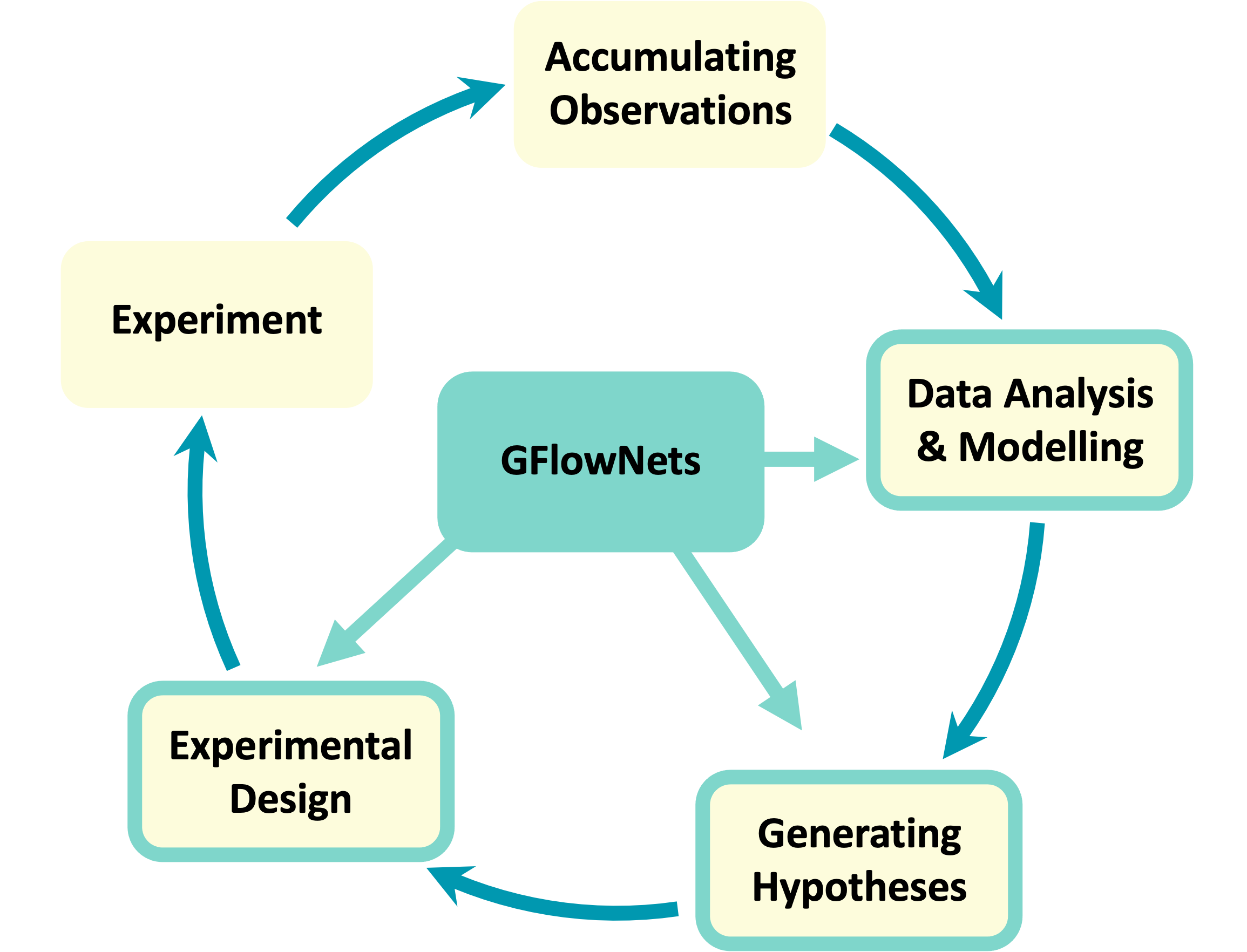}
    \caption{The iterative experimental loop of scientific discovery: observations and data accumulated from past experiments are analyzed and used to generate new hypotheses, and in turn new experiments that will yield new data to continue to cycle. Highlighted with a blue frame are the steps for which we discuss how GFlowNets can be used.}
    \label{fig:experiment-model-cycle}
\end{figure}

Historically, scientific discovery has been the outcome of either serendipity---such as penicilin and teflon~\citep{ban2022serendipity}---or the rather slow experimental science loop: observations are accumulated from past experiments, which are carefully analyzed by experts who produce new hypotheses and design experiments that will eventually yield new, valuable observations to continue the cycle (see Figure~\ref{fig:experiment-model-cycle}). While this model has well served the progress of science for centuries and will continue to do so in certain domains, the fully manual version of this cycle is too slow for the pressing emergencies of our time. A bottleneck in the cycle occurs when the analysis of data, production of hypotheses and experimental specification are manual. This is further exacerbated when the search space of candidates is dauntingly large, as is the case for drug discovery where there exist $10^{60}$ feasible small molecules, according to estimations~\citep{wayne1996}.

The scale at which scientific experiments can be conducted is rapidly increasing, enabled by advances in robotics, biotechnology and computational capabilities, among others \citep{macleod2020sdl}. For example, we can now easily and cheaply collect high-dimensional images and videos, electron microscopy data or the gene expression of millions of cells. Furthermore, we can also conduct thousands or even millions of experiments in parallel to screen new candidate molecules, experiment with a sequence of reactions, etc. If experimental interventions can be combined, we can sample from a 
combinatorially large space of possible experiments at each step. 
The avenues opened by such large-scale availability of data and compute have been identified as \textit{the fourth paradigm in scientific discovery} \citep{hey2009fourthparadigm}. Nonetheless, our current tools are not enough to truly utilize all the information and resources at our disposal \citep{agrawal2016fourthparadigm}. In this context, the maturation of tailored machine learning (ML) methodologies offers the possibility to not only analyze and make sense of the data, but also to improve the generation of hypotheses and design of experiments, accelerating the experimental science loop.

ML techniques have been employed in all of the main steps of the experimental science loop illustrated in Figure~\ref{fig:experiment-model-cycle}: (a) analyzing and modeling the data accumulated from experiments, (b) characterizing and generating hypotheses compatible with the data, (c) designing the next experiments, and (d) performing the experiments. 
Analysis and modeling of data is a naturally appealing scenario for ML methods, which are typically designed for extracting predictive patterns from large data sets. For instance, machine learning methods have been quite successful in modeling the quantum mechanical properties of small molecules~\citep{stark20223d}. ML approaches have also been studied in the context of designing candidate experiments~\citep{ryan2016review}. A standard example of this is leveraging tools from reinforcement learning (RL) and Bayesian optimization (BO) for searching candidates that optimize (as a reward) some property of the candidate~\citep{angermueller2019model,kim2022deep}. However, as we discuss in detail in Section~\ref{sec:challenges}, existing approaches often lack a principled treatment of the challenges introduced by limited data, uncertainty and underspecification of the objectives in scientific domains.

In this paper, we discuss how a novel machine learning (ML) framework, called Generative Flow Networks~\citep{bengio2021flow,bengio2021gflownet}---or GFlowNets for short---can help in addressing the shortcomings of existing ML approaches in scientific domains. GFlowNets are general purpose \emph{inference machines}, which enable generating samples with a probability proportional to some reward function. As a consequence, GFlowNets have emerged as a potentially transformative tool for scientific discovery, as they can be used to generate hypotheses, design experiments and model the experimental observations, key steps of the experimental science loop (Figure~\ref{fig:experiment-model-cycle})---note that ML-augmented robotics can also be useful in the experimental step, but we do not discuss this here. 

Throughout the paper, we consider the following motivating examples to make the discussion more concrete.

\begin{example}
An important part of fighting growing antimicrobial resistance and emergent infectious diseases is speeding up the discovery of novel small organic molecules (and peptides) that inhibit the action of target bacteria or one to several target proteins. The primary goal is to search the space of molecules (including peptides) for candidates that bind to the target of interest and inhibit or activate its function. The space of molecules is combinatorially large, and the accurate evaluation of the desired activity (e.g. binding affinity) in-vitro or even in-silico is expensive. Additionally, aside from binding to a target, there are several other pharmacological criteria which a molecule needs to satisfy for use as a therapeutic, such as low toxicity to humans, good Absorption, Distribution, Metabolism, and Excretion (``ADME''), and ease of synthesis.
\label{ex:drug_discovery}
\end{example}
\begin{example}
Discovery of novel materials for applications in the generation, storage, and use of clean energy that have better efficiency and rely on sustainable raw materials are critical in aiding efforts to reduce rising global temperatures. The goal here is discovering novel inorganic as well as organic materials, and there are often several specific metrics to optimize simultaneously. A representative example is the development of storage materials for lithium/sodium ions in lithium/sodium ion batteries, where the class of metal oxides (e.g. Figure~\ref{fig:gfn-schematics}) alone already represents a combinatorially large search space. Inside a lithium/sodium ion battery, a good candidate metal oxide for energy storage should show high energy density, together with other metrics such as high capacity retention, low irreversible capacity, and low cost of synthesis.   
\label{ex:material_discovery}
\end{example}
\begin{example}
Modeling the genetic pathways through which diseases progress within humans plays a critical role in our understanding of human biology. These causal models can help us understand the behavior of various interventions such as therapeutics within the complex environment of the human body. The goal is to learn such causal models with the help of targeted interventions. Even the causal structure of a single cell is a major challenge and raises difficult questions to appropriately scale algorithms and combine learning from data with prior knowledge from biology.
\label{ex:causality}
\end{example}

\paragraph{Organization.}
In Section~\ref{sec:background} we discuss existing ML methodology, highlighting the distinctive features of GFlowNets. Next in Section~\ref{sec:gfn} we introduce ideas around amortized inference, and discuss how GFlowNets emerge from these ideas. In Section~\ref{sec:diverse-generation}, we highlight scientific discovery problems where GFlowNets have been applied successfully.
In Section~\ref{sec:posteriors-causal-models}, we discuss how GFlowNets can also be used to represent a distribution over causal models linking multiple random variables of interest and to estimate the posterior distributions of interest along with the marginalized quantities, such as Bayesian posterior densities, that are important to evaluate the information gain from an experiment. To conclude, in Section~\ref{sec:open-challenges} we chart a potential path towards a unified framework for scientific discovery driven by GFlowNets.

\section{Challenges for AI in Scientific Discovery}
\label{sec:challenges}
In the last few decades, ML has enabled remarkable technological advances ranging from agents that can surpass humans at the game of Go~\citep{silver2016mastering} to breakthrough advances in protein folding prediction~\citep{jumper2021highly}. These advances have been enabled, in part, by the availability of extremely large datasets and often of a well-specified objective to be optimized. In many scientific discovery applications, however, the limited available data, the uncertainty intrinsic to measurements and the underspecification of objectives pose serious challenges for leveraging ML approaches. In this section, we discuss the relevance of these challenges and argue how GFlowNets can circumvent them.

\subsection{Limited Data and Uncertainty}
\label{sec:challenges-data}
One critical challenge in leveraging learning-based approaches for scientific discovery is the limited availability of data and the associated uncertainty. Part of the uncertainty is due to unreliable measurements, called \textit{aleatoric uncertainty}, and part is due to having a limited amount of training data, called \textit{epistemic uncertainty}, which is the uncertainty associated with theories or models and their parameters~\citep{KIUREGHIAN2009105}, due to the finite size of datasets. Epistemic uncertainty emerges because multiple theories or models or settings of parameters can be compatible with the given data, and Bayesian posteriors on these can capture the epistemic uncertainty. By design, current state-of-the-art ML approaches rely on access to large data sets to extract useful patterns. But owing to experimental limitations, it can be extremely expensive or impossible to obtain large amounts of accurate and precise data at the scale required my modern ML approaches in many applications of interest. In drug discovery, for example, obtaining experimental binding affinities for ligands with a target protein, at the scale required for ML methods, is often very challenging. Furthermore, the differences in experimental techniques and measurement noise can lead to different binding affinities for the same ligand by orders of magnitude. The same is true in materials science: as an example, in the research for new battery materials, a few thousands of data points are already considered high-throughput~\citep{honrao2021batteries}, and the measurement metrics such as energy density can vary significantly based on minute details of the experimental setup. Thus, it is essential for models to account for the aleatoric and epistemic uncertainty within the context of scientific discovery. 

Additionally, scientific phenomena occurring in nature tend to be complex and often a product of complicated processes. Standard machine learning models that learn a function mapping an input to an output from data can fail to generalize to unseen scenarios where the phenomenon occurs. Incorporating the causal structure of the phenomenon can introduce effective inductive biases that can allow models to generalize to novel scenarios~\citep{scholkopf2021toward}. Whereas a given model will account for uncertainty in outcomes, i.e., aleatoric uncertainty, by modeling Bayesian posteriors we can account for all the models that fit well the data, thus also capturing the epistemic uncertainty, which is what we want to reduce through experiments. As we discuss in Section~\ref{sec:posteriors-causal-models}, GFlowNets can be employed to model the posterior distribution of causal models that fit the data well.

\subsection{Underspecification and Diversity}
\label{sec:challenges-diversity}
ML approaches often assume access to some reward signal or scoring function to evaluate the quality and utility of experimental designs. For instance, to design drug-like molecules, the true objective is to find ligands that specifically inhibit the target protein within the human body. However, this objective can hardly be specified and conveniently quantified as a simple scalar reward. In practice, an estimate of the binding affinity of the molecule with the target protein is used instead as the reward signal to search for candidate molecules. This comes with two caveats: first, the estimation of binding affinity is not immune to systematic and random errors; second, the binding energy {\em alone} cannot account for many of the factors that can influence the effect of the ligand within the human body. For instance, a molecule (or a series of similar molecules) that only optimizes this binding energy may be potentially useless or even harmful in-vivo because it could bind to many other proteins and thus be toxic (more broadly, the molecule could have poor ADME). These aspects make it critical to find diverse hypotheses---in this case, diverse motifs of molecules---to account for the underspecification and uncertainty in the reward signal. Nonetheless, popular approaches to ML-aided scientific discovery, like RL~\citep{angermueller2019model} and Bayesian optimization~\citep{kim2022deep} aim to discover a single maximizer of the the reward signal, not accounting for underspecification of the reward signal itself. Diversity of candidate solutions is also particularly relevant in the evolution of new generations of technologies. For example, before the use of pervoskites (or other new materials) in solar cells, optimization of power conversion efficiency and manufacturing in silicon-based solar cells was yielding diminishing returns, and only the use of radically different materials was able to change this situation. By sampling proportional to the reward, GFlowNets, can mitigate the problem of missing out potentially interesting findings due to underspecification in the target reward, generating a diverse set of high-reward candidate solutions to the discovery problem at hand.

\section{Background}
\label{sec:background}

In this section, we review the relevant fundamental areas in machine learning and set the stage for Section~\ref{sec:gfn} where we discuss how GFlowNets can address the challenges presented in Section~\ref{sec:challenges}.

\paragraph{Preliminaries.} . 
To establish notation, consider 
an example problem from organic synthesis where chemists want to find new and/or efficient synthetic pathways to obtain a desired molecule (e.g. a drug candidate).  

\begin{itemize}
    \item \textbf{Design.} Let $x\in \mathcal{X}$ be the design for an experiment, where $\mathcal{X}$ denotes the design space of all possible experiments. For Example~\ref{ex:drug_discovery} and Example~\ref{ex:material_discovery}, each experimental design, $x$, specifies a candidate molecule, antibody or metal oxide material. $x$ can also represent interventions on specific genes for Example~\ref{ex:causality}, or experimental parameters for a specific procedure (e.g. synthesis conditions). If an experiment is modeled in-silico, it can also include the fidelity of the approximations used, with the associated computational cost.
    \item \textbf{Parameters.} We use $\theta$ to represent the parameters of our mathematical model of the underlying phenomenon of interest. Like the experimental design $x$, $\theta$ can parameterize a wide range of objects. For instance, $\theta$ could represent the parameters of the physical process of supramolecular interactions/protein binding in Example~\ref{ex:drug_discovery}, or a set of parameters describing the model of energy capacity and mechanisms of capacity loss in a battery for Example~\ref{ex:material_discovery}, or a structural causal model describing the causal interaction genetic pathways in Example~\ref{ex:causality}, or more general cases such as parameters of a neural network. 
    \item \textbf{Outcome.} We use $y$ to denote the experimental outcome, for example the yield of a chemical reaction. The outcome, $y$, may also be multi-dimensional, and may include all measurements recorded during the experiment. $y$ can represent the binding affinity to a target, toxicity to humans and synthetic accessibility in Example~\ref{ex:drug_discovery}. For Example~\ref{ex:material_discovery}, $y$ can represent energy capacity and retention loss, as well as materials purity and X-ray diffraction patterns. In Example~\ref{ex:causality}, $y$ can even include images and videos of specifc interventions on a population of cells. The experimental outcomes are often structurally rich but might lack the abstractions necessary for effective modeling. 
    \item \textbf{Dataset.} Finally, we denote the data collected from previous experiments as $\data = \{(x_i, y_i)\}_{i=1}^M$. 
\end{itemize}

We model the outcome $y$ as a consequence of the design $x$. 
The \emph{likelihood}---denoted $p(y\mid \theta, x)$---is parameterized by $\theta$, and is a measure of how likely each experimental outcome, $y$, is to occur, given a particular design, $x$, and model parameters, $\theta$.  This likelihood acts as our abstract model of the underlying phenomenon. If the likelihood is known analytically or can be computed efficiently it is called an 
\emph{explicit model}. For example, we might model the outcome of an experiment with a Gaussian distribution, with mean as some function $f_\theta(x)$, such that $p(y\mid \theta, x) = \mathcal{N}(f_\theta(x); \sigma)$. Alternatively, if the likelihood is intractable, it is called an \emph{implicit model}. Implicit models are common in the context of scientific discovery because we often model processes via simulators that allow us to sample from $p(y\mid \theta, x)$, without being able to evaluate the likelihood.

We assume that the data set of observed variables $\data=\{x_1, \dots, x_n\}$,
is drawn from the joint distribution $p(x,\theta)$ for some $\theta$. 
In order to use that data to update our model of likely outcomes, we need an approach to solving the
\emph{inference} problem: estimating the posterior probability distribution $p(\theta\mid \data)$ over the latent variable given the observed data. This problem appears in various contexts across domains. For instance, given observations of the experimental outcomes, say the binding affinities of several ligands to a particular target, we might be interested in estimating the distributions over parameters in the model that describe the process of binding. In principle this distribution can be estimated using Bayes' rule,

\begin{equation}
    p(\theta\mid\data) = \frac{p(\data\mid\theta)p(\theta)}{p(\data)}.
\end{equation}
In practice, however, computing the posterior exactly is intractable for high-dimensional $\theta$ and $x$. Thus, approximate inference methods have been studied extensively. We briefly summarize two broad classes of methods: Markov Chain Monte Carlo (MCMC) and Variational Inference (VI).

\subsection{Approximate Inference}
\label{sec:approx-inference}

\textbf{MCMC methods}~\citep{andrieu2003introduction} are designed to generate samples from a target distribution, where a correct sampling procedure is not known but the density of the desired distribution is known up to a normalizing constant. They approximate sampling from the desired distribution by constructing a sequence of samples whose asymptotic distribution (as the sequence becomes longer) matches the desired target distribution. At each step of the sequence a new sample is generated by performing a small random perturbation from the previous sample. When trying to sample from a Bayesian posterior $p(\theta\mid \data)$, we can compute the unnormalized form of the posterior as the product of the prior and the likelihood, $p(\theta)p(\data\mid\theta)$. Unfortunately, in high-dimensional problems and problems where the modes of the distribution occupy a tiny relative volume and can be far from each other, MCMC methods can take exponentially more time to properly sample from all the modes (or even just move from one mode to another). Methods to improve the performance of MCMC for sampling from high-dimensional distributions~\citep{torrie1977nonphysical,earl2005parallel,beskos2011hybrid} are limited to certain classes of distributions and do not apply to sampling complex objects such as graphs and sequences which are important in a number of applications in scientific discovery.

\textbf{Variational Inference methods}~\citep{blei2017variational} approach the problem of sampling from the posterior using instead an optimization-based approach, finding a member of family of distributions that is closest to the posterior distribution that we seek. In particular, they search for a distribution $q(\theta)$---by optimizing the values of $q$ or parameters that define it---so as to minimize the reverse Kullback-Liebler (KL) divergence $E_{\theta\sim q}[\log q(\theta) - \log p(\theta, \data)]$. Because this measure of ``closeness'' is a reverse KL, VI methods tend to drop most modes of the true posterior or even focus on just one of them~\citep{minka2005divergence,turner2011two}. 

GFlowNets address this issue of mode dropping that plagues both MCMC and typical VI methods. They are similar to {\em amortized} variational methods (i.e., they learn a parameterization for $q$) but use a different training objective which favors a greater diversity of samples 
by allowing the use of exploration in the space of samples~\citep{malkin2022gflownets} as we discuss in Section~\ref{sec:amortized-inference}. 

\subsection{Experimental Design}
\label{sec:experiment-design}

Experiments are the primary interface for interaction between our abstract models and the complexities of the real world. A key element of scientific methodology has been the careful design of experiments that allow the acquisition of knowledge corresponding to a reliable understanding of the underlying phenomena. 
However, experiments are expensive---either computationally, financially or in time. Therefore, we need methods to design experiments that maximize the amount of information our models learn from each experiment. This task of automated experimental design has been extensively studied in statistics and machine learning. 

The field of experimental design studies the problem of designing ``useful'' experiments effectively. The usefulness of an experiment is defined by a utility function (or reward) $U(\cdot; \data):\mathcal{X}\rightarrow \mathbb{R}$, which may change as a function of the data, $\data$, that we have observed from previous experiments. Given this utility function, we are typically interested in selecting the most useful designs, $x^*$, 
\begin{equation}
    x^* = \arg \max_{x \in \mathcal{X}}\, U(x; \data).
\end{equation}

The process of experimental science is often iterative, as illustrated in Figure~\ref{fig:experiment-model-cycle}. We design an experiment, perform the experiment and observe the experimental outcome, update our model based on the observations and then design the next experiment guided by the updated model. This is referred to as \emph{sequential experimental design}.
The sequential experimental design setting is thus formalized at any iteration $k$ as follows, in terms of the estimated utility of the experiment $x$ considered and the past data $\data_{k-1}$:
\begin{equation}
   x_{k} = \arg \max_{x\in\mathcal{X}} U(x; \data_{k-1})
\end{equation}
where $\data_{k-1} = \{(x_i, y_i)\}_{i=1}^{k-1}$ consists of the designs and outcomes of the experiments performed till iteration $k$. 

Designing utility functions that accurately reflect the value of an experiment while being efficient to compute has been a problem of interest in various communities. Classical work on experimental design relied on the Fisher information matrix to quantify the information about parameters $\theta$ contained in the experimental outcome $y$~\citep{fedorov1972theory,atkinson1992optimum}. 
This measure of information can be efficiently computed in linear models, where the outcome depends linearly on the design~\citep{ryan2016review}. When this relationship is nonlinear,  a variety of methods exist to select among a \textit{version space} of nonlinear functions that are consistent with what what we have observed; see ~\citet{settles2012active} for a survey. 
In scientific discovery, we are not agnostic to the set of functions that could explain our observations: we typically have
 significant 
prior knowledge from the literature and previous experiments that what we can use to weigh the relative likelihood of potential experimental outcomes. Bayesian experimental design, introduced below, provides a principled approach to incorporating these priors into our choice of future experiments.

\subsection{Bayesian Experimental Design}
\label{sec:BOED}

Bayesian Experimental Design (BED) or {Bayesian Optimal Experimental Design} (BOED)~\citep{chaloner1995bayesian,ryan2016review} approaches experimental design by modeling a rational agent that aims to maximize their expected utility of new experiments with respect to prior beliefs. The utility function provides a real-valued score for each potential outcome, $y$, of any potential experimental design, $x$. At each round of experimentation, the agent selects an experimental design\footnote{Sequences of experiments are also possible, but even more computationally involved.}, $x$, that gives the highest expected utility weighted by how likely each outcome is to occur under the agent's prior beliefs (parameterized by $\theta$). 
By specifying the agent's prior belief, we can encode scientific knowledge of known relationships and uncertainties in the observed outcomes, thereby making the procedure more efficient at exploring unknown parts of the experimental design space.

A common choice for the agent's utility function is the \emph{mutual information}~\citep[MI;][]{lindley1956measure} between the experimental  parameters $\theta$ and the outcome $y$ observed upon performing an experiment $y$, given the dataset $\data$ of previous experiments. 

\begin{align}
    U(x;\data) &= I(y;\theta\mid x, \data) \\
    &= \mathbb{E}_{p(y\mid \theta, x)p(\theta\mid \data)}\left[\log \frac{p(\theta\mid y,x, \data)}{p(\theta\mid \data)}\right] \label{eq:ptheta-mi}\\
    &= \mathbb{E}_{p(y\mid \theta, x)p(\theta\mid \data)}\left[\log \frac{p(y\mid \theta, x)}{p(y\mid x, \data)}\right]. \label{eq:py-mi}
\end{align}

MI can be interpreted as how much information can we expect to gather---or equivalently how much we reduce our uncertainty---about some random variable of interest (say the model parameters $\theta$) thanks to the experimental outcome. Equations~\ref{eq:ptheta-mi} and~\ref{eq:py-mi} give two equivalent definitions of MI, but both are intractable in general so we will need to rely on approximations.

Assuming access to the likelihood function, $p(y\mid \theta, x)$, we first need to estimate the various posterior distributions, depending on which of the above formulations of the MI we choose.  
To sample one of the sums, we need to estimate or at least sample from either the unknown and generally intractable posterior over parameters $p(\theta\mid\data)$, or the marginal likelihood $p(y\mid x)$ (where $\theta$ has been summed out). Estimating high-dimensional posteriors can be challenging and marginalizing out $\theta$ in $p(y\mid\theta, x)$ can also be intractable. Additionally, the MI itself involves a high dimensional integral which can be intractable. Consequently, developing efficient estimators to approximate the MI has been one of the central challenges in BED. Estimators of MI have been developed for both implicit and explicit models. The estimators typically leverage tools from approximate inference, including MCMC and VI discussed in Section~\ref{sec:approx-inference}, to approximate the posterior over parameters or the marginal likelihood~\citep{myung2013tutorial,rainforth2018nesting,foster2019variational,kleinegesse2019efficient,kleinegesse2020bayesian,heinrich2020information}, and the likelihood in the case of implicit models~\citep{overstall2020bayesian}.

\paragraph{Towards Amortization}
Conventional approaches discussed above consist of two distinct steps: estimating or approximating the posterior over parameters or marginal likelihood to estimate the MI and then maximizing this estimator of MI to find the optimal experiment. Despite being applied in various contexts in scientific discovery~\citep{drovandi2013sequential, overstall2020bayesian}, each of these steps alone can be computationally expensive. \cite{foster2020unified} introduced a unified stochastic gradient method to combine estimation of MI and the selection of the optimal experiment. They propose jointly optimizing the parameters $\phi$ of the MI estimator and the experiment design $x$ using gradient based methods.

\cite{foster2021deep} formalize the conventional BED approach in terms of a design policy $\pi$, which directly maps a history of experimental data $h_t=[(x_1, y_1), \dots (x_t, y_t)]$ to the next experiment design $x_{t+1}$. Once this policy is trained, the next experiment can be selected directly using the policy, instead of the usual MI estimation and optimization. In essence, the training of the policy \emph{amortizes} the cost of estimating and optimizing the MI. \cite{ivanova2021implicit} extends this to implicit models. 

These are examples of {\em amortized inference}: instead of expensive Monte-Carlo sampling to estimate or maximzes some expected value at run-time, we pre-train a function that directly produces an approximation of the desired quantities. We elaborate on learned amortized inference in Section~\ref{sec:gfn}, with a focus on GFlowNets and a discussion of its advantages over MCMC methods.

Current BED methods have been limited to continuous design spaces. While recent work has considered extensions to incorporate discrete designs~\citep{blau2022optimizing}, they are limited to small problem domains. BED methods in general are hard to scale to larger problem settings.

\subsection{Bayesian Optimization}
\label{sec:bayesopt}
A typical problem encountered in various domains of scientific interest is that of optimizing the value of some expensive to compute black-box function $f$. For instance, consider the task of designing novel molecules to inhibit the activity of a particular target protein. We are interested in searching for a molecule $x^\star$ which minimizes the binding energy of the molecule with the target protein, $f$: 

\begin{equation}
    x^\star=\arg \min_{x \in \mathcal{X}}f(x).
\end{equation}

Further, we only observe noisy measurements $y$ on evaluating $f$, as described by $p(y\mid f(x))$. In the context of the binding energy, each experimental evaluation can be expensive and take weeks to perform in the lab and the observed results are noisy. Consequently the goal here is to discover $x^\star$ with the fewest possible evaluations of $f$. This problem is studied broadly within the framework of \emph{global optimization}~\citep{horst2000introduction}. Solving the global optimization problem for any general function $f$ is NP-hard in discrete spaces, and intractable without special structural constraints (e.g. convexity) on $f$ in the continuous case. Methods for finding approximate solutions to this problem have received significant attention in the literature due to the broad applicability. Among the wide variety of techniques studied, \emph{Bayesian Optimization} (BO)~\citep{mockus1978application,jones1998efficient,garnett_bayesoptbook_2022} is a popular and widely used approach. Bayesian Optimization has been applied extensively to a wide variety of scientific problems~\citep{gonzalez2015bayesian, griffiths2020constrained,moss2020gaussian, shields2021bayesian}. Broadly, Bayesian optimization consists of an iterative process to search for the global optimum in a sample-efficient manner, relying on tools from Bayesian Inference. Algorithm~\ref{algo:bayesopt} provides a general overview of the Bayesian optimization approach. 

\begin{algorithm}[H]
 \textbf{Input}: Oracle $f\sim p(f)$, Initial dataset, $\data_0 = \{(x_i, y_i)\}_{i=1}^M$, Acquisition function $\alpha$\;
 \KwResult{$x_T \approx \arg \max_{x\in \mathcal{X}}f(x)$}
 \For{$i=1\dots T$}{
  Infer surrogate model $p(f\mid\data)$ given dataset $\data$\;
  Acquire $x_i = \arg \max_{x\in\mathcal{X}}\alpha(x, p(f\mid \data))$\;
  Observe $y_i\sim p(y\mid f(x_i))$\;
  Update dataset $\data = \data \cup \{(x_i, y_i)\}$ 
 }
 \caption{Bayesian Optimization}
 \label{algo:bayesopt}
\end{algorithm}
 
The two key ingredients of a Bayesian optimization algorithm are the \emph{surrogate model} $p(f\mid \data)$ which approximates $f$ (and its epistemic uncertainty) and the \emph{acquisition function} $\alpha(x, p(f\mid \data))$ which quantifies the utility of acquiring a point. Gaussian Processes~\citep[GPs;][]{rasmussen2005gaussian} are an appealing choice for the surrogate model owing to simple analytical form for the posterior $p(f\mid \data)$. Consequently, GPs are the default choice in modern BO methods~\citep{balandat2020botorch}. 

From an information theoretic perspective, in each round we are interested in acquiring candidates that maximize the mutual information between the observed value and the global optimum of the function: 
\begin{equation}
    \arg \max_{x\in\mathcal{X}} I(y; x^\star \mid x, \data).
    \label{eq:bo-info}
\end{equation}

This objective resembles the information gain from Section~\ref{sec:BOED}, where the parameter of interest is $\theta = x^\star = \arg \max_{x\in\mathcal{X}}f(x)$. Indeed, BO can be viewed as an instantiation of experimental design with an implicit model $f$, where we are interested in a particular random variable, the location of the maximum value of $f$, rather than all the parameters.

\paragraph{Acquisition Functions} The acquisition function plays a critical role in Bayesian Optimization and over the years various acquisition functions have been proposed. Expected Improvement~\citep{mockus1991bayesian} was one of the earliest acquisition functions. Upper-Confidence Bound~\citep[UCB;][]{srinivas2010gaussian} and Thompson Sampling~\citep[TS;][]{thompson1933likelihood,vakili2021scalable} were inspired by the bandit learning literature. More recently, there have been significant developments in \emph{entropy search} methods which adopt the information theoretic perspective introduced in Equation~\ref{eq:bo-info}. \cite{hennig2012entropy} and \cite{hernandez2014predictive} introduced Entropy Search (ES) and Predictive Entropy Search (PES) as acquisition functions based on Equation~\ref{eq:bo-info}. \cite{hoffman2015output,wang2017max} instead considered the mutual information between the outcome and the \emph{max value} of $f$ rather than the $\arg \max$, resulting in the Max Value Entropy Search (MES) acquisition function:
\begin{equation}
    \arg \max_{x\in\mathcal{X}} I(y; f(x^\star) \mid x, \data).
    \label{eq:bo-mes}
\end{equation}
\cite{moss2021gibbon} further proposed considering a lower-bound on the mutual information resulting in a general purpose information-theoretic acquisition function GIBBON, which is also applicable to various extensions we discuss below.

\paragraph{Extensions}
Inspired by various practical applications, several extensions to the standard Bayesian Optimization setting have been studied. A common scenario is where $f$ can be evaluated on multiple different candidates in parallel. In practice, we can often evaluate multiple candidates with nearly the same cost as a single candidate. For example, phage display can produce libraries of millions of antibodies in one batch. Batch Bayesian Optimization~\cite{gonzalez2016batch,kandasamy2018parallelised} is an extension of BO where in each round we acquire a batch of candidates instead of a single candidate. Additionally, we might have access to oracles with different costs and fidelities to evaluate $f$; for example, to obtain the simulated binding affinity of a molecule to a protein, oracles can include free energy perturbation and molecular docking, where the former is substantially more accurate but the computational cost is orders of magnitude higher.. This setting is studied in Multi-fidelity Bayesian optimization~\citep{picheny2010noisy,kandasamy2017multi,takeno2020multi}. Another important aspect in practical applications is multiple objectives. For example, we are interested in multiple properties such as the drug-likeness, toxicity to humans, and synthesizability in addition to binding energy in the drug discovery setting. Multi-Objective Bayesian Optimization~\citep{hernandez2014predictive,belakaria2019max,daulton2022multi} methods study this problem setup. Recent work has also incorporated physical inductive biases as priors for efficient Bayesian Optimization~\citep{ziatdinov2022hypothesis,mcdannald2022fly}. Finally, while traditional BO methods mainly consider continuous $x$, recent work has enabled BO on discrete spaces~\citep{moss2020boss,swersky2020amortized}. Despite recent progress, BO methods have typically been limited to small problems due to challenges in scaling surrogate models to larger domains. Additionally, as mentioned in Section~\ref{sec:challenges}, as BO is concerned with maximizing or minimizing a function, it can miss out on diversity which is critical for many scientific applications. 

\subsection{Causal Discovery}
\label{sec:causal-discovery}
An important goal of the scientific methodology is understanding the causes and effects of certain phenomena based on prior observations and experiments. Causal Discovery studies the problem of learning the causal structure from data. 

A \emph{Bayesian Network}~\citep{pearl1988bayesiannetworks} is a representation of the joint distribution over $d$ random variables $\{Y_{1}, \ldots, Y_{d}\}$, whose conditional independencies are encoded in a compact graphical way. These random variables correspond to the nodes of a directed acyclic graph (DAG) $G$ that determines the factorization of the joint distribution as
\begin{equation*}
    P(Y_{1}, \ldots, Y_{d}; \theta) = \prod_{i=1}^{d}P(Y_{i}\mid \mathrm{Pa}_{G}(Y_{i}); \theta_{i}),
\end{equation*}
where $\mathrm{Pa}_{G}(Y_{i})$ is the set of parent nodes of $Y_{i}$ in the graph $G$, and $\theta_{i}$ are the parameters of the conditional distribution associated with the random variable $Y_{i}$.

Although in a Bayesian Network the edges connecting the nodes in $G$ only encode associations between random variables, a \emph{causal graphical model} enhances this framework with notions of causality. In a causal graphical model, again represented by a DAG, $G$, over the random variables, any directed edge $Y_{i} \rightarrow Y_{j}$ represents a direct causal influence of $Y_{i}$ on $Y_{j}$. This allows the model to not only represent the joint distribution of the system (i.e., passively observing the system), but also the effects of actively experimenting on it. A causal model specifies the distribution that would be obtained under any intervention. For example, a ``DO-intervention'' sets the value of a variable, ignoring its usual causes. However, because the same causal mechanisms (the conditionals $P(Y_{i}\mid \mathrm{Pa}_{G}(Y_{i}); \theta_{i})$) are shared across all interventions, if the causal mechanisms (i.e. $\theta$) and the graph $G$ have been inferred correctly, a causal model can generalize to distributions never seen during training (i.e., out-of-distribution), corresponding to new interventions.

\subsubsection{Causal Structure Learning}
\label{sec:causal-structure-learning}
The structure $G$ of a causal graphical model is often assumed to be known, where for example the causal relationships are determined using expert knowledge. This allows us to perform a number of tasks using these models, such as \emph{inference} (either probabilistic, or causal), or \emph{learning} the parameters $\theta$ of the causal model from observations of the system.

However in the context of scientific discovery, the objective is precisely to discover causal relationships that may have eluded experts thus far. For example, in the development of a disease, we want to find what factors (e.g. social factors, proteomes, pathogens) are involved. In this situation, we would like to learn the structure of the causal graphical model (or at least part of it) using data, stored in a dataset $\data$. This data could either come from passive observations of the system (called \emph{observational} data, e.g. the statistics of protein expressions in patients), or from active experiments (called \emph{interventional} data, e.g. genome-wide gene perturbation). This problem is known as \emph{causal structure learning}, or \emph{causal discovery} \citep{spirtes2000pc,chickering2002ges,zheng2018notears,ke2019causal,brouillard2020dcdi}.

\subsubsection{Bayesian Causal Discovery}
Similar to how there may be multiple theories explaining the same phenomenon, there may also be multiple models that could explain our observations equally well, even in the limit where we have a very large amount of data. Concretely, this means that many standard structure learning methods would typically choose an arbitrary model, which could lead to undesirable (and potentially harmful) outcomes. For example, if we had a system with only two (correlated) random variables $A$ and $B$, there would be no way in general to distinguish between the two causal models $A \rightarrow B$ and $B \rightarrow A$ using observational data only, even though both models have significantly different causal conclusions. Moreover, in practice, the amount of data available in $\data$ to identify the causal model may be scarce, and this introduces another source of variability: since causal discovery methods only return a single candidate, some theory may be favored only due to the limited evidence. Ideally, we would like to quantify our \emph{epistemic uncertainty} to avoid model misspecification. This can be done using the \emph{Bayesian posterior} over the causal structures $G$, given a dataset $\data$, similar to the description in Sec.~\ref{sec:bayesian-ml}. Using Bayes' rule, the posterior is given by
\begin{equation}
    P(G\mid \data) = \frac{P(\data\mid G)P(G)}{P(\data)}.
    \label{eq:bayes-rule-causal-discovery}
\end{equation}
In the expression above, $P(G)$ represents our \emph{prior belief}, and may encode some a priori knowledge about the structure $G$. For example, we may encourage causal graphs to be sparse, i.e. to limit the number of parents for any node in the graph. While this prior may be designed based on expert knowledge, this usually encodes only soft beliefs about the causal model, and does not represent a single graph $G$ unlike when the structure is assumed to be known. The term $P(\data\mid G)$ is called the \emph{marginal likelihood}, and is defined by integrating over all possible values of the causal mechanisms
\begin{equation}
    P(\data\mid G) = \int_{\Theta}P(\data\mid \theta, G)P(\theta \mid G) d\theta,
    \label{eq:marginal-likelihood-causal-discovery}
\end{equation}
where $P(\data\mid \theta, G)$ is the likelihood of the data under a specific choice of causal structure and mechanisms, and $P(\theta \mid G)$ is a prior distribution over causal mechanisms. The marginal likelihood represents how well the data $\data$ fits a certain hypothesis, given by the causal model $G$, regardless of the choice of the causal mechanisms themselves.

As is typically the case in Bayesian statistics, the difficulty in evaluating Eq.~\ref{eq:bayes-rule-causal-discovery} arises from the marginal evidence $P(\data)$, which is almost always intractable. To circumvent this issue, approximations of the Bayesian posterior are often necessary, for example based on MCMC \citep{madigan1995structuremcmc,friedman2003ordermcmc,giudici2003improvingmcmc,niinimaki2016partialordermcmc,viinikka2020gadget}, bootstraping \citep{friedman1999bootstrap,agrawal2019abcd}, or more recently variational inference \citep{annadani2021vcn,cundy2021bcdnets,wang2022trust,lorch2021dibs,lorch2022avici}.

\subsubsection{Active Learning of Causal Structures} With an approximation of the Bayesian posterior over causal graphs, we can also leverage the tools from Bayesian Experimental Design in Section~\ref{sec:BOED} in order to design interventions on the system that would refine our beliefs about its causal structure \citep{buntine1991theory}. This creates a feedback loop between the estimation of our uncertainty about the causal graph based on data, the decisions about which experiments to perform, and the acquisition of new experimental data (see Figure~\ref{fig:experiment-model-cycle}). \emph{Active causal discovery} can be used to learn the causal structure of either the whole system \citep{tong2001active,murphy2001activelearning,scherrer2021learning,tigas2022interventionswhere}, or part of it \citep{agrawal2019abcd}. Recently, \citet{toth2022abci} proposed a novel framework called \emph{Active Bayesian Causal Inference} (ABCI) to infer not only the causal graph, but also jointly learning the posterior over causal queries of interest.

\section{Generative Flow Networks}
\label{sec:gfn}
We begin by introducing the broader ideas around using neural networks to efficiently learn a mapping from sampled observations to proposed high-dimensional probability distributions and intractable sums, as a general substitute for the popular MCMC-based inference. These ideas lead into GFlowNets, which provide a general framework for amortized inference with neural networks.

\begin{figure}
\centering
    \includegraphics[width=0.9\textwidth]{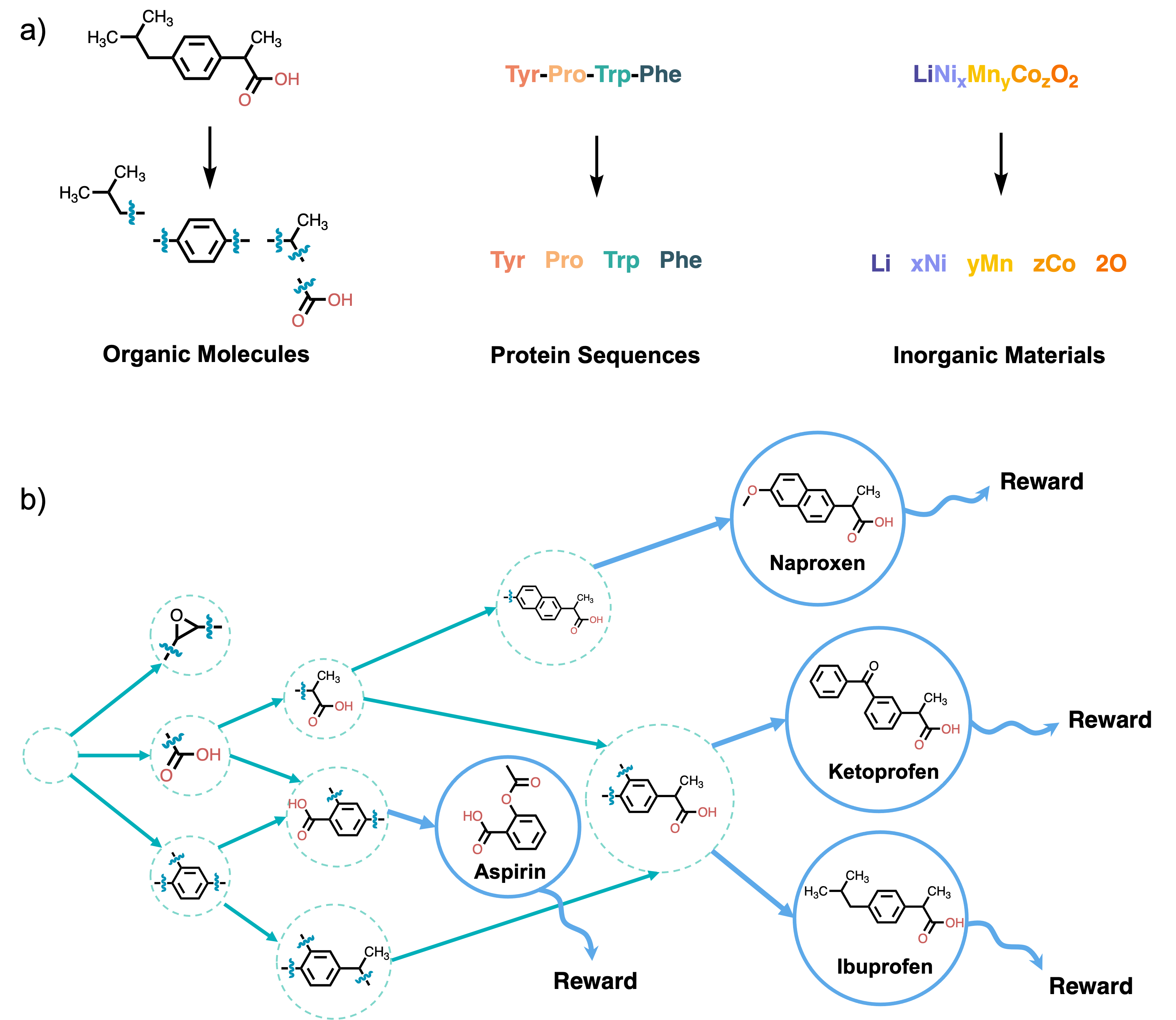}
\caption{Example schematics of a GFlowNet. (a) The objects whose distribution is modeled by a GFlowNet should be compositional, like graphs or sequences, built through a sequence of actions; (b) Representation of the sequences of actions by which a GFlowNet can construct a molecule by composing smaller fragments, as a directed acyclic graph (DAG) whose nodes represent partially constructed molecules, and a reward is provided when the molecule is completed.}
    \label{fig:gfn-schematics}
\end{figure}

\subsection{Learning to Perform Amortized Inference}
\label{sec:amortized-inference}

Let us first look at how neural nets can be used to amortize the inference problem introduced in Section~\ref{sec:approx-inference}, i.e., by being trained to approximately perform the sampling or summing task that is otherwise intractable. Specifically, we consider how an intractable expectation or sum can be transformed into a tractable training task to approximate the desired sum. This is the fundamental principle underlying GFlowNets.

\subsubsection{Simple Mean Squared Error Criterion to Amortize an Intractable Expectation}

Consider a set of intractable expectations that we would like to approximate, for a pair of random variables $x$ and $y$ that can both take an exponential number of values or live in a high-dimensional space:
\begin{equation}
S(x) = \sum_y p(y\mid x) R(x,y)
\end{equation}
 which is then intractable because of the exponential number of terms in the sum. 

If we know how to sample from $p(y\mid x)$, we could, however, train a neural net $\widehat{S}$ with input $x$, stochastic target output $R(x,y)$ and Mean Squared Error (MSE) loss 
\begin{equation}
L(x,y) = \left( \widehat{S}(x) - R(x,y) \right)^2
\end{equation}
where $y \sim p(y\mid x)$, to train the estimator $\widehat{S}$ with parameters $\theta$. When we sample training examples $(x,y)$, the stochastic gradients $\frac{\partial L(x,y)}{\partial  \theta}$  would make $\widehat{S}$ converge to $S$ if it has enough capacity and is trained long enough~\citep{bishop1995neural}.  

For any new $x$, we would then have an amortized estimator $\widehat{S}(x)$ which in one pass through the network would give us an approximation of the intractable sum $S(x)$. We can consider this an efficient alternative to doing a Monte-Carlo approximation 
\begin{equation}
\widehat{S}_{MC}(x) = {\rm mean}_{y \sim p(y\mid x)} R(x,y),
\end{equation}
which would require a potentially large number of samples and computations of $R(x,y)$ for each $x$ at run-time, especially if $p(y\mid x) R(x,y)$ is a rich multimodal function (for which averaging just a few samples of $y$ does not give us a good estimator of the expectation). 

Besides the advantage of faster run-time, a crucial potential advantage of the amortized version is that it could benefit from  {\em generalizable structure} in the product $p(y\mid x) R(x,y)$: if observing a training set of $(x, y, R(x,y))$ triplets can allow us to generalize to new $(x,y)$ pairs, then we may not need to train $\widehat{S}$ with an exponential number of examples before it captures the generalizable structure and provides good answers (i.e., approximates $E_{Y\mid x}[R(x,Y)]$ well) on new $x$'s. This ability to generalize from structure in the data is actually what explains the remarkable success of ML (and in particular of deep learning) in the vast set of modern AI applications.

\begin{figure}
\centering
    \includegraphics[width=0.5\textwidth]{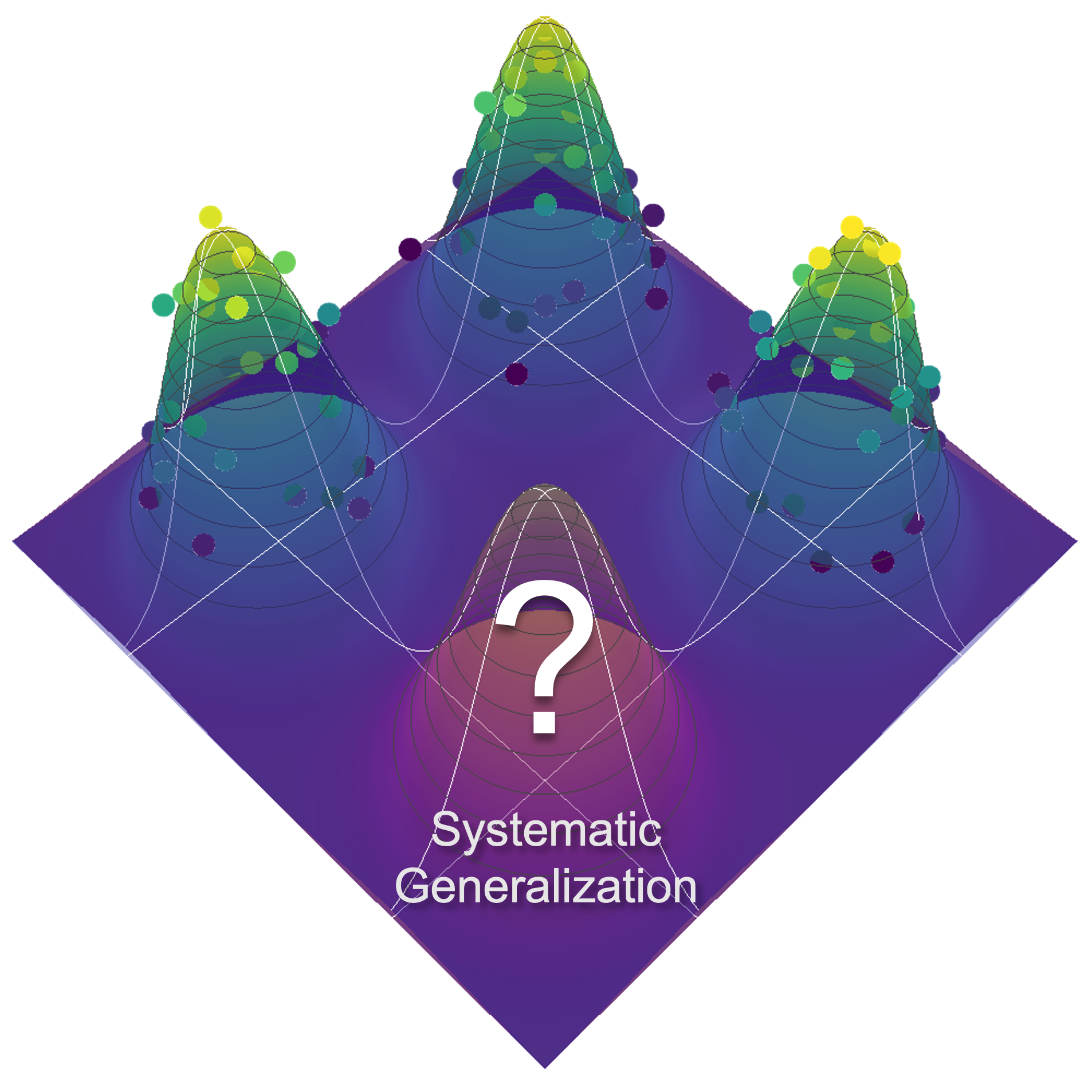}
\caption{An illustration of the feasibility of systematic generalization enabled by ML methods: if we have already discovered the three modes shown of a reward function, a learner that can generalize may guess the presence of a 4th mode, because the first three modes seem to align on a grid. The existence of such generalization structure is why amortized ML samplers can potentially do much better than MCMC samplers.}
    \label{fig:mode-generalization}
\end{figure}

When we do not have a $p(y\mid x)$ that we can sample from easily, we can, in principle, use MCMC methods that form chains of samples of $y$'s whose distribution converge to the desired $p(y\mid x)$, and where the next sample is generally obtained from the previous one by a small stochastic change that favors increases in $p(y\mid x)$. Unfortunately, when the modes of the summand, $p(y\mid x) R(x,y)$, occupy a small volume in the search space (i.e. throwing darts does not find them) and these modes are well-separated (by low-probability regions), especially in high dimension, it tends to take exponential time to mix from one mode to another. However, such an MCMC approach leaves money on the table: the attempts $(x,y,R(x,y))$ contain information that one could use to train an ML model. To the extent that the space is sufficiently structured, such a model could guess where the yet unseen modes might be given the location of the already observed modes, as illustrated in Figure~\ref{fig:mode-generalization}.

\subsubsection{GFlowNet Criterion to Obtain a Sampler and Estimate Intractable Sums}

Let us consider the situation where we do not have a handy $p(y\mid x)$ and our objective is just to approximate a set of intractable sums (for any $x$) 
\begin{equation}
S(x) = \sum_y R(x,y)
\end{equation}
where we have the constraint that $R(x,y) \geq 0$ and $S(x)>0$. This may be useful to estimate a normalization constant for energy-based models or Bayesian posteriors (where $y$ corresponds to learnable parameters and $x$ to observed data). Hence we may also be interested in the sampling policy
\begin{equation}
\pi(y\mid x) = \frac{R(x,y)}{S(x)} \propto R(x,y).
\end{equation}
Now, GFlowNet losses are derived from a set of constraints that we would like to be true: 
\begin{equation} 
\forall (x,y):\quad \pi(y\mid x) S(x) = R(x,y). 
\label{eq:pi-S=R-constraint}
\end{equation}
We can define estimators $\widehat{\pi}$ and $\widehat{S}$ and train them with a loss such as
\begin{equation}
L(x,y) = \left( \widehat{\pi}(y\mid x) \widehat{S}(x) - R(x,y)\right)^2
\end{equation}
or with an interpretation of $R$ as unnormalized probabilities that we want well calibrated in the log-domain, 
\begin{equation}
L(x,y) = \left( \log(\widehat{\pi}(y\mid x) \widehat{S}(x)) - \log R(x,y)\right)^2
\end{equation}
where $(x,y)$ are sampled from a training distribution $\widetilde{p}(x,y)$ that has full support. It can then be shown~\citep{bengio2021flow,bengio2021gflownet} that with $\widehat{\pi}$ and $\widehat{S}$ with enough capacity and trained for long enough they both converge to their desired value: 
\begin{align*} 
\widehat{S}(x) &\rightarrow S(x) \\ \widehat{\pi}(y\mid x) &\rightarrow \pi(y\mid x) \propto R(x,y). 
\end{align*}
This is a crucial property of GFlowNets: \emph{they are trained to sample objects $y$ (given $x$) with probability proportional to a given reward function $R(x,y)$.}

\subsubsection{Marginalizing over Compositional Random Variables}

To make the notion of intractable sum more concrete, it is good to think of $y$ (and potentially $x$ as well) as a compositional object, like a subset of variable-value pairs or a graph. For example, we can construct compositional objects sequentially through a series of steps where a new piece of the compositional object is inserted at each step, such as the molecular fragments composed to form a larger molecule in Fig.~\ref{fig:gfn-schematics}b. The sampling policy $\pi$ then sequentially and stochastically chooses a constructive action at each step, and after each step we get a {\em partially constructed object} $s$ which we call a GFlowNet {\em state}. A sequence of such states and actions forms a GFlowNet {\em trajectory} $\tau$. 
In the basic GFlowNet framework, the actions are stochastic, but the next state is deterministically obtained from the previous state and the action,  because they are not happening in an external environment but are part of the internal computation of a sampler. In addition, the GFlowNet mathematical results, as they currently stand, assume that each step is constructive, i.e., we cannot return to the same partially constructed object $s$ twice. This means that the set of all possible trajectories forms a directed acyclic graph (DAG), illustrated in Figure~\ref{fig:gfn-schematics}. 
Because the transitions are deterministic, we can specify a trajectory $\tau$ with a sequence of states (or, equivalently, an initial state and a sequence of actions). A special ``exit" action is also defined to declare the construction of the object $y$ is completed. The policy $\pi(y\mid x)$ is now specified by a forward transition distribution $P_F(s\mid s')$ which specifies how to generate each constructive step given the previous state, and we are interested in parameterizing and learning this $P_F$.

For instance, consider the molecule graph example illustrated in Figure~\ref{fig:gfn-schematics}.
We can construct a molecule graph sequentially using nodes representing atoms as building blocks. 
Starting from a special empty state, which we denote as $s_0$ (this can be an empty graph, null set or empty sequence or chosen based on the value of a conditioning variable $x$, maybe specifying some desired characteristics of the molecule), the object $y$ can be constructed through a sequence of steps, each consisting of adding a single block $a \in \mathcal{A}$. We assume that the actions are limited to be constructive, and deletion of blocks is not allowed.
At each step, we have a partially-constructed object $s\in \mathcal{S}$, where $\mathcal{S}$ denoted the space of all possible partially-constructed objects and $\mathcal{Y}\subset \mathcal{S}$. Another assumption we make throughout is that these states are Markovian, that is, they incorporate all the information from their history.
This results in a \emph{directed acyclic graph} (DAG) $\mathcal{G}$ which is defined by a tuple $\left(\mathcal{S}, \mathcal{E}\right)$, where the set of nodes corresponds to $\mathcal{S}$, and an edge $s\rightarrow s'\in \mathcal{E}$ indicates that object $s'$ can be constructed by adding a block $a\in \mathcal{A}$ to $s$, $s \xrightarrow[]{a} s'$.
We can define a trajectory as a sequence of steps describing the construction of an object $\tau=\big(s_0 \xrightarrow[]{a_1}s_1\dots\xrightarrow[]{a_n}y\big)$. 
\footnote{Note that there can be multiple trajectories resulting in the same object at the end.} 
Let $R(x,y)$ denote the utility (reward) for a given object $y$ in context $x$. For instance, it can be the binding energy for the ligand molecule $y$ with a given target protein $x$. 

Formally, the training objective is to learn a stochastic policy $\pi$ which sequentially generates an object $y$ with a probability proportional to its reward, i.e., $\pi(y\mid x)\propto R(x,y)$.

\subsubsection{Multiple Parent States}

Besides the sequential nature of the generative process for $y$, an interesting complication is that there may be many ways (in fact exponentially many trajectories) to construct $y$ from some starting point and context $x$. This means that a partially constructed object, i.e., a state $s$, may have multiple parents $s'$ for which an action $a$ exists that leads to $s$. Otherwise (when each state only has one parent), the DAG is a tree, which makes the computation much simpler. But when it is not a tree, it turns out to be convenient to consider and parameterize a backward transition probability function $P_B(s' \mid  s)$ which is consistent with that DAG and the associated forward transition probabilities $P_F$. The constraint in Eq.~\ref{eq:pi-S=R-constraint} can be reformulated in several ways, in particular what is called the detailed balance constraint: 
\begin{equation} 
\forall (s,s'):\quad P_F(s\mid s') F(s') = P_B(s'\mid s) F(s) \label{eq:detailed-balance-condition} \end{equation}
where $F(s)$ is called the flow at state $s$ and plays a role similar to $S(x)$ above, i.e., it is an intractable sum, and there is a starting state $s_0=x$ from which a trajectory is initiated, as well as a constraint that the flow into a terminal state $s=y$ equals $R(x,y)$. Similarly to the simpler case  above, this can be turned into a training loss that we want to minimize, but now over all $(x,\tau)$ pairs or over all $(s, s')$ pairs. As a consequence of satisfying the detailed balance constraint at all $(s,s')$ pairs, the initial flow becomes equal to the normalizing constant~\citep{bengio2021flow}: 
\begin{equation}
F(s_0)=S(x) = \sum_y R(x,y).
\end{equation}

\subsubsection{Learning Objectives}
\label{sec:learning-objectives}
In practice, we would like to approximate $P_F(\cdot\mid\cdot;\theta), P_B(\cdot\mid\cdot;\theta)$, and $F(\cdot; \theta)$ with learnable parameters $\theta$, and we want to choose those parameters to satisfy as well as possible the detailed balance constraint $\forall s, s' \in \mathcal{S}$. The detailed balance constraint can be converted to the following loss to learn the parameters $\theta$:
\begin{equation}
    \mathcal{L}_{DB}(s,s';\theta) = \left(\log \frac{P_F(s'\mid s; \theta) F(s';\theta)}{P_B(s\mid s';\theta)F(s;\theta)} \right)^2.
    \label{eq:db_loss}
\end{equation}
Several alternative learning objectives for GFlowNets have been proposed, especially for longer trajectories to sample the object $y$~\citep{malkin2022trajectory,madan2022learning}. Trajectory Balance~\citep[][TB]{malkin2022trajectory} is a prominent learning objective for training GFlowNets. Contrary to the detailed balance objective, which considers constraints on pairs of states, trajectory balance jointly applies the detailed balance constraint over entire trajectories. A learnable parameter $Z$ is introduced which at convergence is equal to the desired sum. The trajectory balance loss over a trajectory $\tau$ is defined as follows:

\begin{equation}
    \mathcal{L}_{TB}(\tau;\theta) = \left(\log \frac{Z_\theta\prod_{s\rightarrow s' \in \tau}P_F(s'\mid s;\theta)}{R(x,y)\prod_{s\rightarrow s' \in \tau}P_B(s\mid s';\theta)} \right)^2.
    \label{eq:tb_loss}
\end{equation}

Algorithm~\ref{algo:gfn_train} illustrates the typical approach for training GFlowNets, by sampling trajectories from a sampling policy $\hat{P}_F$ which is typically a tempered $P_F$ or a mixture of $P_F$ with a uniform policy to enable exploration, and optimizing the loss induced by the learning objective with respect to $\theta$, with stochastic gradient descent\footnote{Code implementing various GFlowNet learning objectives on simple synthetic domains: \href{https://github.com/saleml/gfn}{{saleml/gfn}}}. As a result of this procedure, we learn a $P_F$ with the objective that the marginal likelihood of a trajectory terminating at a terminal state $x$, denoted as $\pi(x)$ become proportional to the reward $R(x)$. The learnable objects $P_F, P_B, F$ are typically parameterized by neural networks. These neural networks must have appropriate inductive biases depending upon the type of objects we are constructing. These neural networks must also have enough capacity to model the underlying distribution. In Section~\ref{sec:diverse-generation} and Section~\ref{sec:posteriors-causal-models} we discuss specific cases of leveraging GFlowNets for problems of molecule generation and causal modeling respectively.

\begin{algorithm}[H]
 \textbf{Input}: Reward function $R(x,y)$, Learning objective or training loss $\mathcal{L}$ \;
 \textbf{Initialize}: $F(\cdot;\theta), P_F(\cdot\mid\cdot;\theta), P_B(\cdot\mid\cdot;\theta)$ for DB and $Z_\theta, P_F(\cdot\mid\cdot;\theta), P_B(\cdot\mid\cdot;\theta)$ for TB\;
 \For{$i=1\dots N$}{
  Sample trajectory $\tau$ from the sampling policy $\hat{P}_F(\cdot\mid\cdot;\theta)$ \;
  Compute the training loss $\mathcal{L}$\;
  Update parameters $\theta$ with stochastic gradient descent\; 
 }
 \KwResult{Policy $\pi(x) \propto R(x)$}
 \caption{General recipe for training GFlowNets}
 \label{algo:gfn_train}
\end{algorithm}

\subsubsection{Implications for Bayesian ML}
\label{sec:bayesian-ml}

Let us consider the special case where $y=\theta$ is a latent parameter, i.e., in a Bayesian setting, and $x=\data$ is the available data. Then we can define the reward function 
\begin{equation}
R(\data,\theta) = P(\theta) P(\data\mid\theta)
\label{eq:reward-bayesian}
\end{equation}
from the parameter prior $P(\theta)$ (how plausible are these parameters a priori) and the data likelihood $P(\data\mid \theta)$ (how well does this choice of parameters fit the data). Training a GFlowNet provides us with an approximate sampler for the posterior over parameters 
(the policy $\pi(\theta\mid \data)$) given data as well as an estimator of the normalizing constant of the Bayesian posterior,  through the learned initial flow $S(\data)$. Hence, we have used amortization to turn a tractable function (prior times likelihood, as a function of $\theta$) into estimators of these generally intractable quantities. We get a fast sampler for the posterior with no need for a Markov chain going through a large number of candidate samples. With that sampler, we can generate many independent samples $\theta\mid \data$ (possibly in parallel) that are likely to visit the larger modes of the posterior (where the reward is larger).

To make computations more ML-friendly (especially for large datasets) while training the GFlowNet, we can note that the GFlowNet squared loss objectives naturally lend themselves to the case where the reward or log-reward is stochastic and is an unbiased estimator of the true reward. For example, we can typically decompose the overall dataset log-likelihood $\log P(\data\mid \theta)$ into a sum of per-example or per-minibatch terms, and we can introduce a multiplicative correction to account for the prior $P(\theta)$: 
\begin{align*} 
\log \widehat{R}(Z,\theta) &= \log P(\theta) + |\data| \log P(Z\mid \theta) \\ E_Z[\log \widehat{R}(Z,\theta)] &= \log P(\theta) + \sum_{Z \in \data} \log P(Z \mid  \theta) = \log R(\data,\theta) 
\end{align*}
where the expectation over $Z$ is just a sum over the $Z$'s in $\data$.
This makes it possible to train the GFlowNet posterior estimator using stochastic gradient descent on single examples or minibatches, which is the state-of-the-art to train deep nets.

\subsubsection{Why GFlowNets?}
\label{sec:why-gfn}
Let us look at how GFlowNets differ from other related conceptual frameworks: 
\begin{itemize}
    \item Markov Chain Monte-Carlo: GFlowNets do not construct Markov chains with semantics like those in MCMC methods. GFlowNets and MCMC approaches differ fundamentally: with MCMC approaches, the chain is irreducible (every state is reachable from every other state), and the stationary distribution of this chain is of interest. In order to generate samples from this stationary distribution, we need to run long chains (ideally infinitely long ones). In GFlowNets, however, the chains only need every state to be accessible from the initial state, and what we have is a bounded sequence of stochastic transitions. The expensive stochastic "search" (to reach low energy configurations) normally performed by MCMC is replaced by the training phase of the GFlowNet, using the principle of amortized inference, so that during inference a sample can be generated in a single short trajectory by the policy. To exemplify, if we were to run an MCMC to generate samples of desirable molecules, we would probably have a molecule at every state, and we would have actions that can stochastically transform one molecule into a nearby one~\citep{xie2021mars}, whereas a typical way to sample a desirable molecule with GFlowNets is constructive, where we start from an empty object (which is not really a molecule) and we sequentially and stochastically add molecular fragments: the state corresponds to these intermediate objects, which may or may not correspond to a well-formed molecule~\citep{bengio2021flow}.  That being said, the overall objective is the same: turn a given energy function (or unnormalized distribution) into a generative procedure for obtaining samples from that distribution. Both MCMC and GFlowNets can also be used to marginalize, i.e., compute intractable sums, although again in different ways. MCMC turn the exact sum into a Monte-Carlo approximation of it while GFlowNets perform amortized inference, i.e., they train a neural net whose output will, after training, approximate the sum. This becomes useful when we have more than one marginalization to do, say given a context, and thus the neural net can take that context as input and rapidly produce an estimator of the intractable sum as output.
    \item Reinforcement Learning: GFlowNets learn policies to sample trajectories that land in a terminal state with probability proportional to the reward of the terminal state rather than trajectories that maximize the expected reward, as in standard deep reinforcement learning. As shown by~\citet{bengio2021flow,jain2022biological}, this results in a diversity of samples which is important when the reward function is an imperfect proxy for the property that we actually care about: it avoids putting all our eggs in the wrong basket.
    
    \item Deep generative Models: Traditional generative models in deep learning such as variational auto-encoders or VAEs~\citep{Kingma2014AutoEncodingVB,JimenezRezende2014StochasticBA} or GANs~\citep{goodfellow2014generative} require positive samples to model the distribution of interest, whereas GFlowNets are trained from a reward function.
    \item Variational Inference: variational inference trains an approximate sampler (and the corresponding density) so as to reduce the forward KL-divergence (the evidence lower bound or ELBO) with a given distribution function (playing the same role as the reward). This requires on-policy training (sampling from the learned sampler), which has a tendency for focusing on a single mode rather than find a diversity of modes. Instead (see~\citet{malkin2022gflownets} for details), GFlowNet objectives enable off-policy training without requiring the high-variance importance sampling correction necessary with the ELBO.
\end{itemize}

To summarize, GFlowNets shine in problems with the following properties:
\begin{itemize}
    \item It is possible to define or learn a non-negative reward function which will specify from what distribution the GFlowNet should sample.
    \item The reward function of interest is highly multimodal. This emphasizes the advantage of GFlowNets in terms of diversity of samples. If the reward function was unimodal, existing RL or variational inference methods (which tend to focus on a single mode) could be used instead.
    \item It is advantageous to sample sequentially, e.g., there is compositional structure that can be exploited by sequential generation.
\end{itemize}

\paragraph{Current Limitations.}
\label{sec:gfn-limit}
Until recently, GFlowNets have been limited to sampling from distributions over discrete objects (e.g. graphs). Recent work by ~\citet{lahlou2023continous} presents a theoretical framework for extending GFlowNets to sample from distributions over continuous spaces. Leveraging this framework for sampling from distributions over high-dimensional continuous or mixed (discrete and continuous) spaces remains an open problem. Another potential limitation, shared with other reinforcement learning methods, is that effective credit assignment over very long trajectories (for generating large objects, such as proteins) is more difficult. \citet{pan2023local} take initial steps to tackle this problem, proposing a way to assign partial rewards earlier in the generated trajectory which results in more effective credit assignment. Another open question is that of the best policy for sampling training trajectories for a GFlowNet. Existing theoretical results from \citet{bengio2021gflownet} assume that the policy sampling trajectories should have full support over the space of trajectories, but designing this policy (beyond the heuristics discussed in Section~\ref{sec:learning-objectives}) for sample-efficient learning is an open problem. 

\subsection{Diverse Candidate Generation}
\label{sec:diverse-generation}

A fundamental problem in chemistry is the synthesis of novel chemical structures (e.g. molecules) that satisfy some criteria. As alluded to in Section~\ref{sec:amortized-inference}, generation of molecules to optimize for a particular chemical property is an appealing use case for GFlowNets, because GFlowNets will tend to generate a diverse set of molecules optimizing that property. An important problem in the context of drug discovery, which we introduced in Example~\ref{ex:drug_discovery} is to discover molecules that bind to a particular target, potentially inhibiting the target in the process. From the computational design perspective, molecular docking simulations can give scoring functions to approximately evaluate proposed molecules. More recently, graph neural networks which approximate the binding energy~\citep{zhang2020molecular} are used to approximate docking as they are much faster. As discussed in Section~\ref{sec:challenges} as these scoring functions serve as approximations to the underlying process, it is important to generate \emph{diverse} candidates for downstream applications, to avoid putting all our eggs in the same basket. 

\cite{bengio2021flow} leverage GFlowNets for the problem of diverse molecule generation\footnote{Code for molecule generation \href{https://github.com/recursionpharma/gflownet}{recursionpharma/gflownet}}. Soluble epoxide hydrolase (sEH) in the 4JNC configuration is studied as a target in the paper. It is a useful target as it plays a role in certain respiratory and heart diseases~\citep{chiamvimonvat2007soluble,imig2009soluble}. Autodock Vina~\citep{trott2010autodock} was used for docking the generated molecules to evaluate the binding energy. Docking each molecule with Autodock Vina can be quite slow and takes several minutes to run, making it prohibitively expensive to train a policy directly directly using it as a reward. Instead, the authors rely on a graph neural network, trained using a data set of docking scores for $300,000$ molecules, as the reward for training the policy. The molecules are generated using fragments, as illustrated in Figure~\ref{fig:gfn-schematics}. At each step, the policy picks a fragment from a library to add to the partially constructed molecule, and choose where to place that fragment. The library of fragments is derived from the Zinc database~\citep{irwin2005zinc}. 

The molecule design problem possesses all the key properties discussed in Section~\ref{sec:why-gfn} for GFlowNets to be effective - (a) there is compositional structure in generation as molecules are built using subgraphs with unique chemical properties (b) the reward function is an approximation of what we really care about, as the docking score and its approximation by a neural network (which has epistemic uncertainty associated with it due to finite training) are approximations of the underlying phenomenon of a molecule binding to a target and inhibiting it, and (c) the reward is multi-modal since there can be multiple motifs of molecules that bind well to a given target. 

\begin{figure}
     \centering
     \begin{subfigure}[b]{0.58\textwidth}
         \centering
         \includegraphics[width=\textwidth]{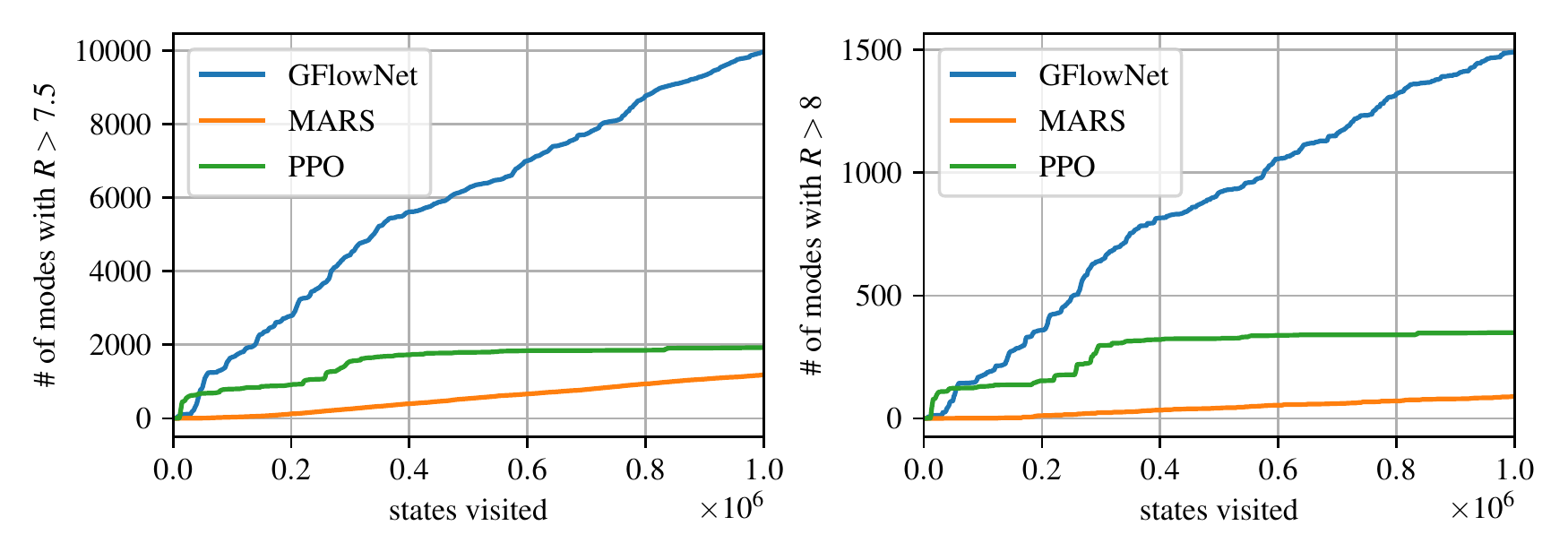}
         \caption{Diversity in generated molecules}
         \label{fig:diversity}
     \end{subfigure}
     \hfill
     \begin{subfigure}[b]{0.35\textwidth}
         \centering
         \includegraphics[width=\textwidth]{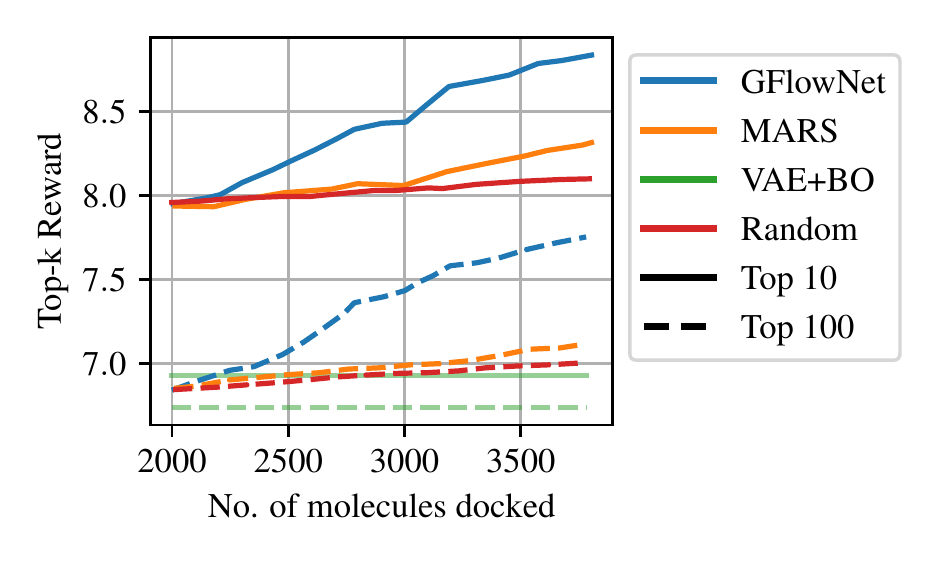}
         \caption{Active Learning}
         \label{fig:al}
     \end{subfigure}
     \caption{(a) Molecules generated with GFlowNets cover significantly more modes of the reward distribution, resulting in diverse high reward molecules. (b) Acquiring molecules generated with GFlowNets results in significant improvements over the starting pool of molecules. Figures taken from~\cite{bengio2021flow} with permission.}
    \label{fig:mol-results-diversity}
\end{figure}

\cite{bengio2021flow} showed that GFlowNets result in substantial improvements over existing methods on this molecule generation task. In particular, as shown in Figure~\ref{fig:diversity}, GFlowNets discover significantly more modes of the reward function (i.e. many different molecules that have high predicted docking score) relative to other reinforcement learning (PPO) and MCMC (MARS) approaches. Sampling proportional to the reward results in high reward and diverse samples. Though it is important to note that while using GFlowNets results in significant improvements in the diversity of generated samples, they do not always lead to the highest scoring candidates, because there is a natural trade-off between diversity and reward. \cite{nica2022evaluating} introduce metrics to study the ability of GFlowNets to explore novel regions in molecular space.

Further, \cite{bengio2021flow} also consider an active learning setup, starting with a data set of $2000$ molecules. In each round, a surrogate model is trained on the data set. This surrogate model is used as the reward for the GFlowNet. Next, molecules are generated with the GFlowNet policy, evaluated with docking, and added to the data set for the next round. Using GFlowNet to acquire the batches of molecules results in significant improvements in the reward over the initial data set, shown in Figure~\ref{fig:al}, demonstrating the potential of GFlowNets to accelerate large-scale virtual screenings. 

From a practical perspective, we are often interested in multiple objectives rather than a single one. For instance, in the context of drug discovery, an ideal drug candidate should specifically inhibit the target but also be synthesizable in large quantities, soluble, and and harmless to humans; alternatively, in material science, to have an efficient solar cell means the optimization of current, voltage, and fill factor. Typically, there are very few candidates which simultaneously satisfy all the objectives, which might even be conflicting with each other (e.g. in solar cells, photocurrent can increase with a photoactive material with lower bandgap, but the voltage decreases). Instead, there exists a set of candidates with the optimal trade-offs between the objectives where further optimizing an objective is impossible without compromising another, i.e. the Pareto front. Moreover, as with the single objective case, diversity is still critical in the multi-objective case. Multi-Objective GFlowNets~\citep[MOGFNs;]{jain2022multi} extend GFlowNets to tackle multi-objective optimization problems. Building upon scalarization approaches in multi-objective optimization~\citep{ehrgott2005multicriteria}, MOGFNs decompose the multi-objective optimization problem into a family of sub-problems which can solved simultaneously. This family of sub-problems is modeled simultaneously with a reward conditional GFlowNet~\citep{bengio2021gflownet}. MOGFNs demonstrate state-of-the-art performance on a variety of small molecule generation and protein design tasks. MOGFNs consistently generate diverse pareto-optimal candidates. For instance, MOGFNs are able to generate molecules that bind to the sEH target, while achieving a high synthesizability and QED score. 

The active learning setting is further explored by \cite{jain2022biological}\footnote{Code for active learning with biological sequences: \href{https://github.com/MJ10/BioSeq-GFN-AL}{mj10/BioSeq-GFN-AL} and \href{https://github.com/alexhernandezgarcia/gflownet}{alexhernandezgarcia/gflownet}}. Incorporating ideas from Bayesian Optimization discussed in Section~\ref{sec:bayesopt}, GFlowNet-AL~\citep{jain2022biological} incorporates information about the epistemic uncertainty of the surrogate model in the reward for the GFlowNet with an acquisition function. This epistemic uncertainty helps in guiding the GFlowNet to optimize the promising less explored regions in the state space. As such, instead of maximizing the acquisition function in Algorithm~\ref{algo:bayesopt}, \cite{jain2022biological} propose sampling proportional to the acquisition function. Equipped with information about the epistemic uncertainty in the reward and other improvements, GFlowNet-AL outperforms various existing methods on a variety of biological sequence design tasks, including generation of peptide sequences with antimicrobial properties. The state space $\mathcal{S}$ consists of partially constructed sequences with each action being the addition of a token from a vocabulary (e.g. a residue from a set of amino acids) to the end of the current partial sequence. Candidate sequences generated by GFlowNets are significantly more diverse and have high rewards. The diversity of generated candidates demonstrates the potential of GFlowNets to accelerate the process of discovering novel antibiotics to tackle the growing and highly concerning phenomenon of antimicrobial resistance~\citep{Oneill-AMR-2014}.

These initial empirical successes demonstrate the potential of GFlowNets to make a significant impact in improving experimental design for a wide variety of scientific problems.

\subsection{Modeling Posteriors over Causal Models}
\label{sec:posteriors-causal-models}

\subsubsection{Bayesian Causal Discovery with GFlowNets}
\label{sec:dag-gflownet}
\begin{figure}[t]
    \centering
    \includegraphics[width=0.8\linewidth]{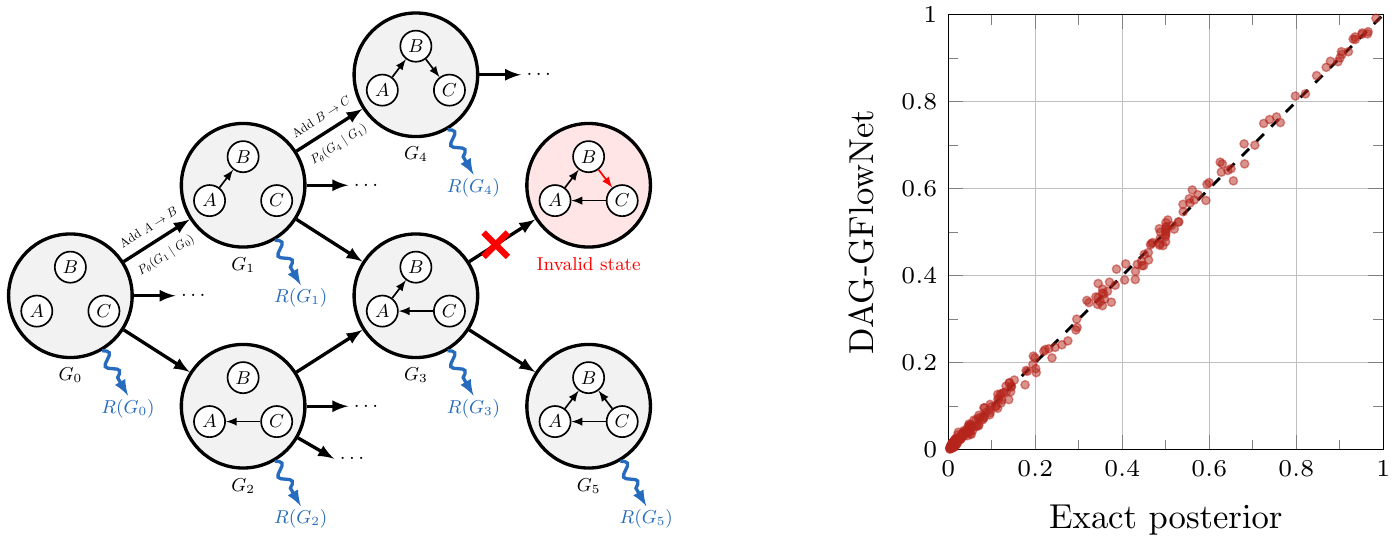}
    \caption{GFlowNet for Bayesian Causal Discovery. (Left) The structure of the GFlowNet introduced in \citep{deleu2022bayesian}, where directed acyclic graphs (DAGs) are constructed one edge at a time. Each DAG is associated with a reward $R(G)$. (Right) Comparison between the edge marginals computed using the exact posterior distribution, and approximated using the GFlowNet; the GFlowNet is capable of accurately approximating the exact posterior $P(G\mid \data)$. Used with permission from Tristan Deleu.}
    \label{fig:dag-gflownet}
\end{figure}

As we have seen in Section~\ref{sec:bayesian-ml}, GFlowNets offer a general solution to approximate Bayesian posteriors, like the one in Eq.~\ref{eq:bayes-rule-causal-discovery}. GFlowNets are all the more adapted to the problem of Bayesian causal discovery that causal structures $G$, represented as a DAG, are compositional objects. \citet{deleu2022bayesian}\footnote{Code for modeling posterior over causal models: \href{https://github.com/tristandeleu/jax-dag-gflownet}{tristandeleu/jax-dag-gflownet}} used this observation to introduce a GFlowNet whose states are DAGs, and where some graph $G$ is created sequentially by adding one edge at a time, starting from the completely disconnected graph over $d$ nodes; the structure of this GFlowNet is shown in Figure~\ref{fig:dag-gflownet} (left). Similar to Eq.~\ref{eq:reward-bayesian}, for a fixed dataset $\data$, the reward function of the GFlowNet is defined as $R(G) = P(\data\mid G)P(G)$. By constraining the set of valid actions at every state, the edges are added in such a way that they will never introduce a cycle, which guarantees that graphs remain acyclic at every stage of the construction. Therefore, all the states of the GFlowNet are valid causal structures. \citet{deleu2022bayesian} leveraged this property and showed that such a GFlowNet may be trained using a modification of the detailed balance loss \citep[][see also Eq.~\ref{eq:detailed-balance-condition}]{bengio2021gflownet}, specifically adapted to the case where all the states are terminating.

To evaluate the reward function $R(G)$ though, one needs to evaluate the marginal likelihood $P(\data\mid G)$ in Eq.~\ref{eq:marginal-likelihood-causal-discovery}, which is in general intractable \citep{heckerman1995bde}. \citet{deleu2022bayesian} experimented only with models such as multinomial-Dirichlet (for discrete data) and linear-Gaussian (for continuous data), for which the marginal likelihood may be computed efficiently in closed form. Alternatively though, instead of approximating the (marginal) Bayesian posterior $P(G\mid \data)$ over structures only, we could approximate the posterior $P(\theta, G\mid \data)$ over both the causal structures $G$ and the causal mechanisms $\theta$ \citep{nishikawa2022bayesian}, to avoid the intractable integration in Eq.~\ref{eq:marginal-likelihood-causal-discovery}. 

\paragraph{Beyond DAGs.}
Work on causal discovery focuses primarily on the causal graphical model framework introduced in Section~\ref{sec:causal-discovery} that assumes a DAG structure, but acyclicity is indeed an assumption that may not hold in certain domains. For instance, in Gene Regulatory Networks, there are feedback loops between multiple genes interacting with one another~\citep{freimer2022systematic}. Nevertheless, when we consider the temporally unfolded cyclic graph, i.e., the dynamics, we are back to a DAG. Some recent work has studied the problem of learning the structure of non-acyclic causal models~\citep{hyttinen2012learning,lorch2022amortized,sethuraman2023nodags}.
The GFlowNet approach discussed above naturally extends to cases where the causal graph might be cyclic. The masking mechanism introduced by ~\citet{deleu2022bayesian} prevents cycles from being introduced at every step where an edge is being added, ensuring the generated graphs are DAGs. If we remove this additional constraint and allow the GFlowNet to introduce cycles in the generated graph, then it would approximate the posterior distribution over cyclic causal models which fit the given observations.

\subsubsection{Bayesian Posteriors for Scientific Discovery}
\label{sec:bayesian-posterior-scientific-discovery}
\paragraph{Posterior Predictive} Once the structure of the causal graphical model is known, we can use the model to perform inference, i.e. answering (possibly causal) questions about the system of interest, 
in the context of different interventions (which we can interpret as setting some variables, i.e., designing an experiment). If we have information about the epistemic uncertainty, through the Bayesian posterior $P(G\mid \data)$, we can even go one step further and average out the predictions made with all possible causal models, weighted using the posterior distribution. Concretely, given a new observation $y$ in the context of an intervention $x$ this corresponds to evaluating
\begin{equation}
    P(y \mid x , \data) = \sum_{G}P(y\mid x, G, \data)P(G\mid \data).
    \label{eq:posterior-predictive-causal-discovery}
\end{equation}
This is called the \emph{posterior predictive}, or \emph{Bayesian model averaging} \citep{madigan1996bma,hoeting1999bma}. The advantage of Eq.~\ref{eq:posterior-predictive-causal-discovery} over making predictions using a single causal model is that multiple concurrent theories may now participate in those predictions. In this way, we avoid using only theory, which may be incorrect, and we obtain a more conservative answer, thus avoiding catastrophic outcomes due to a single theory (say a particular causal graph $G$) being confidently wrong.

\paragraph{Amortized Posterior Predictive} Instead of performing a Monte-Carlo average to estimate $P(y \mid x, \data)$ for each candidate $x$ as per Eq.~\ref{eq:posterior-predictive-causal-discovery} (which may be fairly expensive if we want to train a policy that is trained by considering a large number of possible $x$'s), we can use a neural network $g_{\phi}(x,y)$ to amortize that calculation. This can be done by training $g$ over $(x,y, G)$ triplets with squared loss
\begin{equation}
\mathcal{L}(\phi) = \mathbb{E}_{P(G\mid \data)}\Big[\mathbb{E}_{P(y\mid x, G, \data)}\big[\big(g_{\phi}(x,y) - P(y\mid x, G, \data)\big)^2\big]\Big].
\end{equation}
Other amortization approaches are possible. For example, using the GFlowNet framework, a policy $Q(y_i \mid  x_i, y_1^{i-1}, x_1^{i-1})$ can be trained to first sample one outcome $y_i$ at a time, given the input experiment specification $x_i$ and the previous experimental results $y_1^{i-1}$ of the previous experiments $x_1^{i-1}$. The GFlowNet constraint to satisfy is that 
\begin{align}
    P(\data,G)&=Q(\data,G) \\
    P(G)P(\data\mid G) &= Q(\data) Q(G\mid \data) \\
    P(G) \prod_{i=1}^{|\data|} P(y_i \mid  x_i,  G) &= Q(G\mid \data) \prod_{i=1}^{|\data|} Q(y_i \mid  x_i, y_1^{i-1}, x_1^{i-1})
\end{align}
where the trained posterior predictive takes explicitly a partial dataset ($x_1^{i-1},y_1^{i-1}$) as input, similarly to neural processes~\citep{garnelo2018conditional} and $Q(G\mid \data)$ is the GFlowNet causal graph sampler as described above, 
except that we allow interventions (different choices of $x$) in the data.

\paragraph{Interpretability} Bayesian posteriors over causal models also provide a natural tool for interpretability, since they encode the belief that a causal structure fits the observed data. By inspecting which causal structures contain a certain edge, we can obtain a belief that certain causal relationships between two random variables exist \citep{eaton2007bayesian}. This is called the \emph{edge marginal} distribution:
\begin{equation}
    P(Y_{i} \rightarrow Y_{j} \in G\mid \data) = \sum_{G}\mathbbm{1}(Y_{i} \rightarrow Y_{j} \in G)P(G\mid \data)
    \label{eq:edge-marginal-causal-discovery}
\end{equation}
Fig.~\ref{fig:dag-gflownet} (right) shows a comparison of the edge marginals computed with the posterior approximation returned by DAG-GFlowNet \citep{deleu2022bayesian} against the exact edge marginals, highlighting the capacity of GFlowNets to accurately approximate the posterior over graphs $P(G\mid \data)$. Note that this kind of comparison is typically limited to small problems where the true posterior $P(G\mid \data)$ may be computed efficiently in closed-form, and in general one may be concerned with the calibration of these estimated marginals \citep{lorch2022avici}.

\section{Towards a Unified Framework for Scientific Discovery with GFlowNets}
\label{sec:open-challenges}
In this paper, we have introduced GFlowNets as a tool for modeling and for experimental design in the context of the scientific discovery loop (Fig.~\ref{fig:experiment-model-cycle}). We have summarized how they have been used and could be further used on both fronts. In this concluding section, we outline research directions based on this early work and aimed at providing scientists with a powerful ML-based framework 
applicable when it is possible to iteratively generate informative experimental data.

\subsection{Exploiting Amortized Causal Bayesian Modeling for Defining the Utility of an Experiment}

An appealing theoretical framework for defining the objective of an experiment is that of information gain introduced in Section~\ref{sec:experiment-design}: \emph{``how much information about a random variable of interest can we expect to gain through the experiment?''} A good policy for experimental design should propose experiments with a high value of this information gain as a reward. Note that this framework is broadly applicable to a wide range of interactive learning domains, such as reinforcement learning and active learning, encapsulating 
 the fundamental problem of exploration. In general, the decision to perform an experiment may not simply be based on the number of bits of information gained but also on the risks and costs involved~\citep{zheng2020sequential}. We can however incorporate the notion of a cost or budget by considering the \emph{information gain per unit of cost} incurred as the reward. 

Information gain can be measured in principle by the mutual information between the outcome of an experiment (a random variable since the experiment has not taken place yet) and the variable of interest (about which we seek to gain information), given the experimental specification and any other knowledge (including data) we may already have. In the simplest and purely unsupervised knowledge-seeking scenario, the variable of interest may be the causal model explaining the outcomes of experiments. In a more targeted scenario, for example in drug discovery, it would be the set of molecules that have certain desirable characteristics (e.g., affinity with a target protein is above a threshold and toxicity is below a threshold and synthesis cost is below a threshold). Let $Y$ be the experimental outcome, $x$ be the experiment specification, and $V$ the variable of interest about which we seek to gain information. The information-theoretic utility for our experimental design would then be defined as 
\begin{align}
\label{eq:general-MI}
I(Y;V\mid x,\data) &= \sum_{y,v} P(y,v\mid x,\data) \log \frac{P(y,v\mid x,\data)}{P(y\mid x,\data)P(v\mid x,\data)} \\
&= \sum_{y,v} P(y,v\mid x,\data) \log \frac{P(y\mid v,x,\data)}{P(y\mid x,\data)} \label{eq:mi-amortized}
\end{align}
where $\data$ is our dataset of prior experimental results $\{(x,y)\}$ and any other constraint we want to exploit to condition the probabilities. With $V$ typically being a much higher-dimensional object than $Y$, Eq.~\ref{eq:mi-amortized} tends to be more practical numerically. To evaluate the expression in Eq.~\ref{eq:mi-amortized} we can leverage the ideas of amortization and GFlowNets to estimate (a) the numerator and denominator probabilities $P(y\mid v,x,\data), P(y\mid x,\data)$ (b) a sampler for the joint $P(y,v\mid x,\data)$, and (c) an estimator $\hat{I}$ of the MI itself, as a function of $x$. In the case where where the variable of interest are the parameters of some underlying process $v=\theta$ (as introduced in Section~\ref{sec:experiment-design}), we can follow the amortization approach outlined in Section \ref{sec:bayesian-posterior-scientific-discovery} to estimate the posterior predictive $P(y\mid x,\data)$ in the denominator.
On the other hand, the likelihood in the numerator $P(y\mid\theta,x,\data)=P(y\mid\theta, x)$ is available in the case of explicit models, and can be approximated in the case of implicit models as discussed in Section~\ref{sec:BOED}. Next to learn a sampler for the joint $P(y,\theta\mid x,\data)$ we first approximate samples from the posterior over parameters $Q(\theta\mid \data)$ following Section~\ref{sec:bayesian-ml}. By combining the samples from the posterior $Q(\theta\mid\data)$ and the likelihood $P(y\mid x, \theta)$ we can learn a sampler $Q(y,\theta\mid x,\data)=Q(\theta\mid \data)P(y\mid x,\theta)$ to approximate the joint $P(y, \theta\mid x, \data)$. 

As for the estimator of MI itself, one possibility is to train a neural network $\hat{I}(x)$ to amortize the expected value over $(\theta,y)$ given $(x,\data)$ using the samples from the joint $Q(y, \theta\mid x, \data)$ and the estimators for the probabilities in the log-prob ratio from with a squared loss 
\begin{equation}
\left( \hat{I}(x) - \log \frac{P(y\mid x,\theta)}{Q(y\mid x,\data)} \right)^2
\end{equation}
where $x$ is sampled from a dataset or a generative model of inputs, $\theta\sim Q(\theta\mid \data)$ and $y\sim P(y\mid x,\theta)$. With enough capacity and training time, $Q$ converges to $P$ and $\hat{I}(x)$ converges to $I(Y;\theta\mid x,\data)$. What is particularly interesting if $x$ is in a high-dimensional space is that it generally won’t be necessary to see more than one value of $y$ and $\theta$ for each value of $x$ in order to train $\hat{I}$, as usual in supervised learning. This can work if it is possible for the learning procedure for $\hat{I}$ to generalize from the $(\theta,x,y)$ triplets used to train it.

\subsection{Additional Open Challenges}

\subsubsection{Modeling and Causality}

One open challenge on the modeling side is to leverage GFlowNets to model Bayesian posteriors beyond causal models. \cite{liu2022gflowout} take an initial step in this direction, using GFlowNets to model the posterior over dropout masks in a neural network. Moreover, in the context of causal models, many challenges remain to extend the work done by~\cite{deleu2022bayesian}. This includes (a) accommodating larger causal graphs efficiently (b) making it possible to handle unobserved causal variables by also learning how the raw inputs (e.g., images) may be related to the causal variables 
(c) learning how experimental choices relate to interventions on the causal variable when this is not known perfectly a priori.

\subsubsection{Experimental design}

Our discussion has focused primarily on the case where we are interested in information gain about the parameter $\theta$ to drive knowledge acquisition in the experimental design loop. As we discussed in the previous section, it is possible to incorporate any random variable $V$ (Eq.~\ref{eq:general-MI}) that we can learn to model and sample. An interesting case is one where the random variable $V$ is an extremum, e.g., the top molecular candidates, for some task, as in Eq.~\ref{eq:bo-mes}. 
Furthermore, in many practical experimental settings, we have access to measurements of varying fidelities as the outcome of our experiments (e.g., computer simulations with varying accuracy-computational cost trade-offs). Such a multi-fidelity setting is discussed above (Sec.~\ref{sec:bayesopt}) but needs to be incorporated within the GFlowNet framework. Another important practical scenario also introduced in the same section is the existence of multiple objectives, with early work to incorporate that in the GFlowNet framework by~\citet{jain2022multi}.

Finally, as introduced in Section~\ref{sec:causal-discovery} these experimental design tools could be integrated within the causal discovery framework, by focusing the knowledge acquisition on the causal graph itself, an object of great value to scientists from an interpretability point of view. This could be achieved by using the graph itself as the target variable $V$ (or a part of it), to drive the experimental design to accelerate the discovery of the causal structure.

\section*{Acknowledgements}
The authors thank DreamFold and Maksym Korablyov for important help, suggestions and feedback, as well as funding from CIFAR, Recursion, IVADO, NSERC, IBM, Intel, and the Quebec government.

\bibliographystyle{unsrtnat}
\bibliography{main}

\end{document}